\documentclass[10pt,twocolumn,letterpaper]{article}

\usepackage[accsupp]{axessibility}  

\usepackage{soul}
\usepackage{svg}

\usepackage[accsupp]{axessibility} 

\usepackage[final]{changes}

\def\etal{et~al.\_}			  

\DeclareMathAlphabet{\altmathcal}{OMS}{cmsy}{m}{n}
\DeclareMathAlphabet{\mathbfit}{OT1}{ptm}{bx}{it}

\newlength\paramargin
\newlength\figmargin

\newlength\secmargin
\newlength\figcapmargin
\newlength\tabcapmargin

\newcommand {\first}[1]{{\color{red}\textbf{#1}}}
\newcommand {\second}[1]{{\color{blue}\underline{#1}}}

\newcommand{\topic}[1]
{
\vspace{1mm}\noindent\textbf{#1}
}

\newcommand{\secref}[1]{Section~\ref{sec:#1}}
\newcommand{\figref}[1]{Figure~\ref{fig:#1}} 
\newcommand{\tabref}[1]{Table~\ref{tab:#1}}

\long\def\ignorethis#1{}

\newbox\jsavebox%

\makeatletter
\newcommand{\providelength}[1]{%
  \@ifundefined{\expandafter\@gobble\string#1}
   {
    \typeout{\string\providelength: making new length \string#1}%
    \newlength{#1}%
   }
   {
    \sdaau@checkforlength{#1}%
   }%
}

\newcommand{\sdaau@checkforlength}[1]{%
  \edef\sdaau@temp{\expandafter\sdaau@getfive\meaning#1TTTTT$}%
  \ifx\sdaau@temp\sdaau@skipstring
    \typeout{\string\providelength: \string#1 already a length}%
  \else
    \@latex@error
      {\string#1 illegal in \string\providelength}
      {\string#1 is defined, but not with \string\newlength}%
  \fi
}
\def\sdaau@getfive#1#2#3#4#5#6${#1#2#3#4#5}
\edef\sdaau@skipstring{\string\skip}
\makeatother

\usepackage[pagenumbers]{cvpr} 

\usepackage{graphicx}
\usepackage{amsmath}
\usepackage{amssymb}
\usepackage{booktabs}
\usepackage{makecell}

\usepackage[pagebackref,breaklinks,colorlinks]{hyperref}

\usepackage[capitalize]{cleveref}
\crefname{section}{Sec.}{Secs.}
\Crefname{section}{Section}{Sections}
\Crefname{table}{Table}{Tables}
\crefname{table}{Tab.}{Tabs.}

\def\cvprPaperID{1648} 
\def\confName{CVPR}
\def\confYear{2023}

\begin{document}

\title{Robust Dynamic Radiance Fields}



\author{
Yu-Lun Liu$^{2}$\footnotemark[1]\quad
Chen Gao$^{1}$\quad
Andreas Meuleman$^{3}$\addtocounter{footnote}{-1}\addtocounter{Hfootnote}{-1}\footnotemark[1]\quad
Hung-Yu Tseng$^{1}$\quad
Ayush Saraf$^{1}$\\
Changil Kim$^{1}$\quad
Yung-Yu Chuang$^{2}$\quad
Johannes Kopf$^{1}$\quad
Jia-Bin Huang$^{1,4}$
\\
$^{1}$Meta \quad
$^{2}$National Taiwan University \quad
$^{3}$KAIST \quad
$^{4}$University of Maryland, College Park\\
{\url{https://robust-dynrf.github.io/}}
}

\twocolumn[{
\renewcommand\twocolumn[1][]{#1}
\maketitle
\begin{center}
    \centering
    \resizebox{\textwidth}{!}{%
    \includegraphics[height=1.0\textwidth]{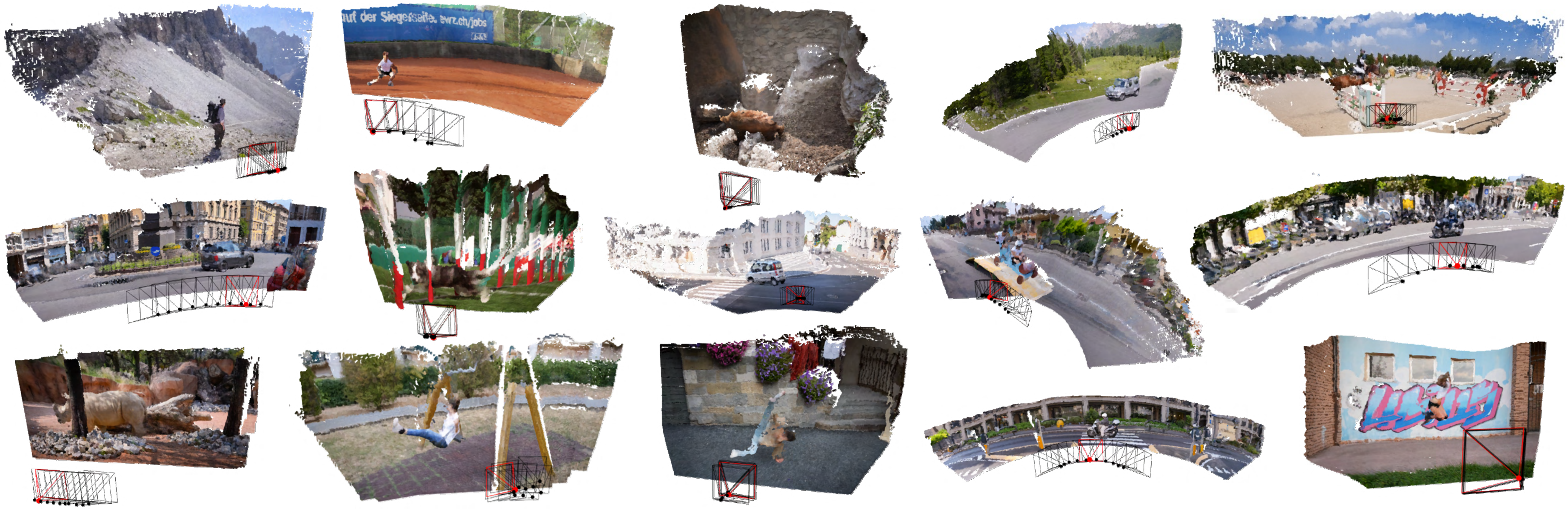}
    }
\captionof{figure}{\textbf{Robust space-time synthesis from dynamic monocular videos.} 
Our method takes a casually captured video as input and reconstructs the camera trajectory and dynamic radiance fields. 
Conventional SfM system such as COLMAP fails to recover camera poses even when using ground truth motion masks.
As a result, existing dynamic radiance field methods that require accurate pose estimation do not work on these challenging dynamic scenes. 
Our work tackles this \emph{robustness} problem and showcases high-fidelity dynamic view synthesis results on a wide variety of videos.
    }
    \label{fig:teaser}
\end{center}
}]

\renewcommand{\thefootnote}{\fnsymbol{footnote}}
\footnotetext[1]{This work was done while Yu-Lun and Andreas were interns at Meta.}

\maketitle


\begin{abstract}
\vspace{-4mm}
Dynamic radiance field reconstruction methods aim to model the time-varying structure and appearance of a dynamic scene. 
Existing methods, however, assume that accurate camera poses can be reliably estimated by Structure from Motion (SfM) algorithms. 
These methods, thus, are unreliable as SfM algorithms often fail or produce erroneous poses on challenging videos with highly dynamic objects, poorly textured surfaces, and rotating camera motion. 
We address this \emph{robustness} issue by jointly estimating the static and dynamic radiance fields along with the camera parameters (poses and focal length). 
We demonstrate the robustness of our approach via extensive quantitative and qualitative experiments. 
Our results show favorable performance over the state-of-the-art dynamic view synthesis methods.
\vspace{-4mm}
\end{abstract}

\section{Introduction}
\label{sec:introduction}


Videos capture and preserve memorable moments of our lives.
However, when watching regular videos, viewers observe the scene from fixed viewpoints and cannot interactively navigate the scene afterward.
Dynamic view synthesis techniques aim to create photorealistic novel views of dynamic scenes from arbitrary camera angles and points of view. 
These systems are essential for innovative applications such as video stabilization~\cite{kopf2014first,liu2021hybrid}, virtual reality~\cite{collet2015high,broxton2020immersive}, and view interpolation~\cite{chen1993view,zitnick2004high}, which enable free-viewpoint videos and let users interact with the video sequence. 
It facilitates downstream applications like virtual reality, virtual 3D teleportation, and 3D replays of live professional sports events.


Dynamic view synthesis systems typically rely on expensive and laborious setups, such as fixed multi-camera capture rigs~\cite{carranza2003free,zitnick2004high,collet2015high,orts2016holoportation,broxton2020immersive}, which require simultaneous capture from multiple cameras. However, recent advancements have enabled the generation of dynamic novel views from a single stereo or RGB camera, previously limited to human performance capture~\cite{dou2016fusion4d,habermann2019livecap} or small animals~\cite{sinha2022common}. While some methods can handle unstructured video input~\cite{ballan2010unstructured,bansal20204d}, they typically require precise camera poses estimated via SfM systems. Nonetheless, there have been many recent dynamic view synthesis methods for unstructured videos~\cite{gao2021dynamic,li2021neural,park2021nerfies,park2021hypernerf,xian2021space,pumarola2021d,gao2022monocular,tretschk2021non} and new methods based on deformable fields~\cite{fang2022fast}.
However, these techniques require precise camera poses typically estimated via SfM systems such as COLMAP~\cite{schoenberger2016sfm} (bottom left of~\tabref{intro}).

However, SfM systems are not robust to many issues, such as noisy images from low-light conditions, motion blur caused by users, or dynamic objects in the scene, such as people, cars, and animals. 
The robustness problem of the SfM systems causes the existing dynamic view synthesis methods to be fragile and impractical for many challenging videos. 
Recently, several NeRF-based methods~\cite{wang2021nerf,lin2021barf,jeong2021self,rosinol2022nerf} have proposed jointly optimizing the camera poses with the scene geometry. 
Nevertheless, these methods can only handle strictly static scenes (top right of~\tabref{intro}).


We introduce RoDynRF, an algorithm for reconstructing dynamic radiance fields from casual videos. Unlike existing approaches, we do not require accurate camera poses as input. Our method optimizes camera poses and two radiance fields, modeling static and dynamic elements. Our approach includes a coarse-to-fine strategy and epipolar geometry to exclude moving pixels, deformation fields, time-dependent appearance models, and regularization losses for improved consistency. We evaluate the algorithm on multiple datasets, including Sintel~\cite{butler2012naturalistic}, Dynamic View Synthesis~\cite{yoon2020novel}, iPhone~\cite{gao2022monocular}, and DAVIS~\cite{perazzi2016benchmark}, and show visual comparisons with existing methods.

\begin{table}[t]
    \caption{
    \textbf{Categorization of view synthesis methods.}
    }
\vspace{-3mm}
    \label{tab:intro}
    \centering
    \resizebox{1.0\columnwidth}{!} 
    {
    \begin{tabular}{c|c|c}
    \toprule
     & \emph{Known} camera poses & \emph{Unknown} camera poses \\
     \midrule
    \raisebox{0.0\normalbaselineskip}[0pt][0pt]{\rotatebox[origin=c]{90}{\makecell{\emph{Static} \\ scene}}} & 
    \makecell{NeRF~\cite{mildenhall2020nerf}, SVS~\cite{riegler2021stable}, NeRF++~\cite{zhang2020nerf++}, \\ Mip-NeRF~\cite{barron2021mip}, Mip-NeRF 360~\cite{barron2022mip}, DirectVoxGO~\cite{sun2022direct}, \\ Plenoxels~\cite{fridovich2022plenoxels}, Instant-ngp~\cite{muller2022instant}, TensoRF~\cite{chen2022tensorf}} & 
    \makecell{NeRF - -~\cite{wang2021nerf}, BARF~\cite{lin2021barf}, \\ SC-NeRF~\cite{jeong2021self}, \\ NeRF-SLAM~\cite{rosinol2022nerf} } \\
    \midrule
    \raisebox{0.0\normalbaselineskip}[0pt][0pt]{\rotatebox[origin=c]{90}{\makecell{\emph{Dynamic} \\ scene}}} & 
    \makecell{NV~\cite{lombardi2019neural}, D-NeRF~\cite{pumarola2021d}, NR-NeRF~\cite{tretschk2021non}, \\ NSFF~\cite{li2021neural}, DynamicNeRF~\cite{gao2021dynamic}, Nerfies~\cite{park2021nerfies}, \\ HyperNeRF~\cite{park2021hypernerf}, TiNeuVox~\cite{fang2022fast}, T-NeRF~\cite{gao2022monocular}} & \textbf{Ours} \\
    \bottomrule
    \end{tabular}
    }
\vspace{-3mm}
\end{table}

We summarize our core contributions as follows:
\begin{itemize}
\item We present a space-time synthesis algorithm from a dynamic monocular video that does \emph{not} require known camera poses and camera intrinsics as input. 
\item Our proposed careful architecture designs and auxiliary losses improve the robustness of camera pose estimation and dynamic radiance field reconstruction.
\item Quantitative and qualitative evaluations demonstrate the robustness of our method over other state-of-the-art methods on several challenging datasets that typical SfM systems fail to estimate camera poses. 
\end{itemize}

\section{Related Work}
\label{sec:related_work}
\begin{figure*}[]
\centering
\includegraphics[width=1.0\textwidth]{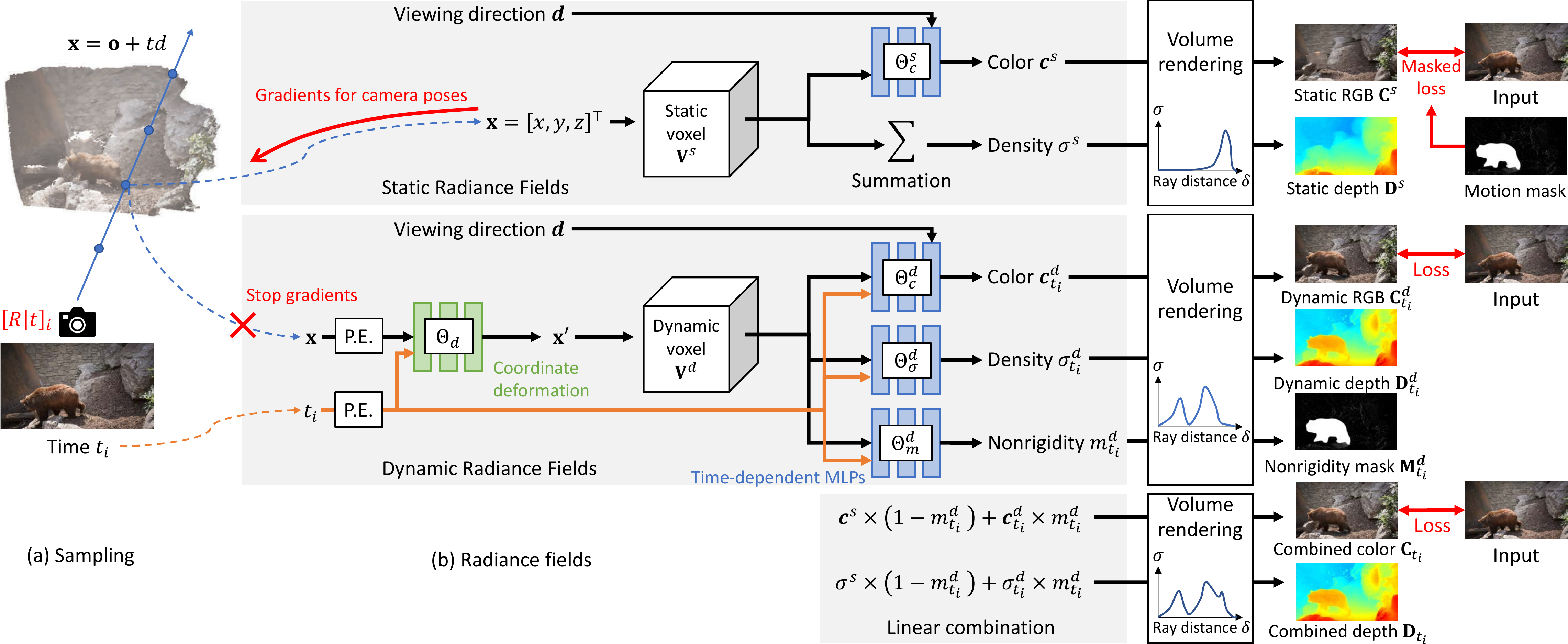}
\vspace{-2mm}
\caption{
\textbf{Overall framework.} 
We model the dynamic scene with static and dynamic radiance fields. 
The static radiance fields take both the sampled coordinates $(x, y, z)$ and the viewing direction $\mathbf{d}$ as input and predict the density $\sigma^s$ and color $\mathbf{c}^s$. 
Note that the density of the static part is invariant to time and viewing direction, therefore, we use summation of the queried features as the density (instead of using an MLP). 
We only compute the losses over the static regions.
The computed gradients backpropagate not only to the static voxel field and MLPs but also to the camera parameters. 
The dynamic radiance fields take the sampled coordinates and the time $t$ to obtain the \emph{deformed coordinates} $(x', y', z')$ in the canonical space.
Then we query the features using these deformed coordinates from the dynamic voxel fields and pass the features along with the time index to a time-dependent shallow MLPs to get the color $\mathbf{c}^d$, density $\sigma^d$, and nonrigidity $m^d$ of the dynamic part. 
Finally, after the volume rendering, we can obtain the RGB images $\mathbf{C}^{\left \{ s, d \right \}}$ and the depth maps $\mathbf{D}^{\left \{ s, d \right \}}$ from the static and dynamic parts along with a nonrigidity mask $\mathbf{M}^d$. Finally, we calculate the per-frame reconstruction loss. 
Note that here we only include per-frame losses. 
}
\label{fig:overview}
\end{figure*}

\topic{Static view synthesis.}
Many view synthesis techniques construct specific scene geometry from images captured at various positions~\cite{buehler2001unstructured} and use local warps~\cite{chaurasia2013depth} to synthesize high-quality novel views of a scene. 
Approaches to light field rendering use implicit scene geometry to create photorealistic novel views, but they require densely captured images~\cite{levoy1996light,gortler1996lumigraph}. 
By using soft 3D reconstruction~\cite{penner2017soft}, learning-based dense depth maps~\cite{flynn2016deepstereo}, multiplane images (MPIs)~\cite{flynn2019deepview,choi2019extreme,srinivasan2019pushing}, additional learned deep features~\cite{hedman2018deep,riegler2020free}, or voxel-based implicit scene representations~\cite{sitzmann2019scene}, several earlier work attempt to use proxy scene geometry to enhance rendering quality.

Recent methods implicitly model the scene as a continuous neural radiance field (NeRF)~\cite{mildenhall2020nerf,zhang2020nerf++,barron2021mip} with multilayer perceptrons (MLPs). 
However, NeRF requires days of training time to represent a scene.  
Therefore, recent methods~\cite{fridovich2022plenoxels,muller2022instant,sun2022direct,chen2022tensorf} replace the implicit MLPs with explicit voxels and significantly improve the training speed.

Several approaches synthesize novel views from a single RGB input image. These methods often fill up holes in the disoccluded regions and predict depth~\cite{niklaus20193d,li2019learning}, additionally learned features~\cite{wiles2020synsin}, multiplane images~\cite{tucker2020single}, and layered depth images~\cite{shih20203d,kopf2020one}. 
Although these techniques have produced excellent view synthesis results, they can only handle static scenes. 
Our approach performs view synthesis of \emph{dynamic scenes} from a single monocular video, in contrast to existing view synthesis techniques focusing on static scenes.

\topic{Dynamic view synthesis.}
By focusing on human bodies~\cite{weng2022humannerf}, using RGBD data~\cite{dou2016fusion4d}, reconstructing sparse geometry~\cite{park20103d}, or producing minimal stereoscopic disparity transitions between input views~\cite{ballan2010unstructured}, many techniques reconstruct and synthesize novel views from non-rigid dynamic scenes. 
Other techniques break down dynamic scenes into piece-wise rigid parts using hand-crafted priors~\cite{kumar2017monocular,russell2014video}. 
Many systems cannot handle scenes with complicated geometry and instead require multi-view and time-synchronized videos as input to provide interactive view manipulation~\cite{zitnick2004high,bansal20204d,broxton2020immersive,lin2021deep}. 
Yoon~\etal~\cite{yoon2020novel} used depth from single-view and multi-view stereo to synthesize novel views of dynamic scenes from a single video using explicit depth-based 3D warping.

A recent line of work extends NeRF to handle dynamic scenes~\cite{xian2021space,li2021neural,tretschk2021non,park2021nerfies,pumarola2021d,gao2021dynamic,park2021hypernerf,fang2022fast}. 
Although these space-time synthesis results are impressive, these techniques rely on precise camera pose input.
Consequently, these techniques are not applicable to challenging scenes where COLMAP~\cite{schoenberger2016sfm} or current SfM systems fail. 
Our approach, in contrast, can handle complex dynamic scenarios \emph{without known camera poses}.


\topic{Visual odometry and camera pose estimation.}
From a collection of images, visual odometry estimates the 3D camera poses~\cite{engel2014lsd,engel2017direct,mur2015orb,mur2017orb,newcombe2011dtam}. 
These techniques mainly fall into two categories: direct methods that maximize photometric consistency~\cite{zhou2017unsupervised,yin2018geonet} and feature-based methods that rely on manually created or learned features~\cite{mur2015orb,mur2017orb,schonberger2016structure}. 
Self-supervised image reconstruction losses have recently been used in learning-based systems to tackle visual odometry~\cite{zhou2017unsupervised,godard2019digging,kopf2021robust,teed2021droid,zhao2022particlesfm,ballester2021dot,bescos2018dynaslam,yang2019cubeslam,yu2018ds,zhang2020vdo}. 
Estimating camera poses from casually captured videos remains challenging. 
NeRF-based techniques have been proposed to combine neural 3D representation and camera poses for optimization~\cite{lin2021barf,wang2021nerf,jeong2021self,rosinol2022nerf}, although they are limited to static sequences. 
In contrast to the visual odometry techniques outlined above, our system simultaneously optimizes camera poses and models \emph{dynamic objects} models.


\begin{figure*}[]
\centering
    \resizebox{1.0\textwidth}{!} 
    {
    \begin{tabular}{cc}
    \includegraphics[height=0.3\textwidth]{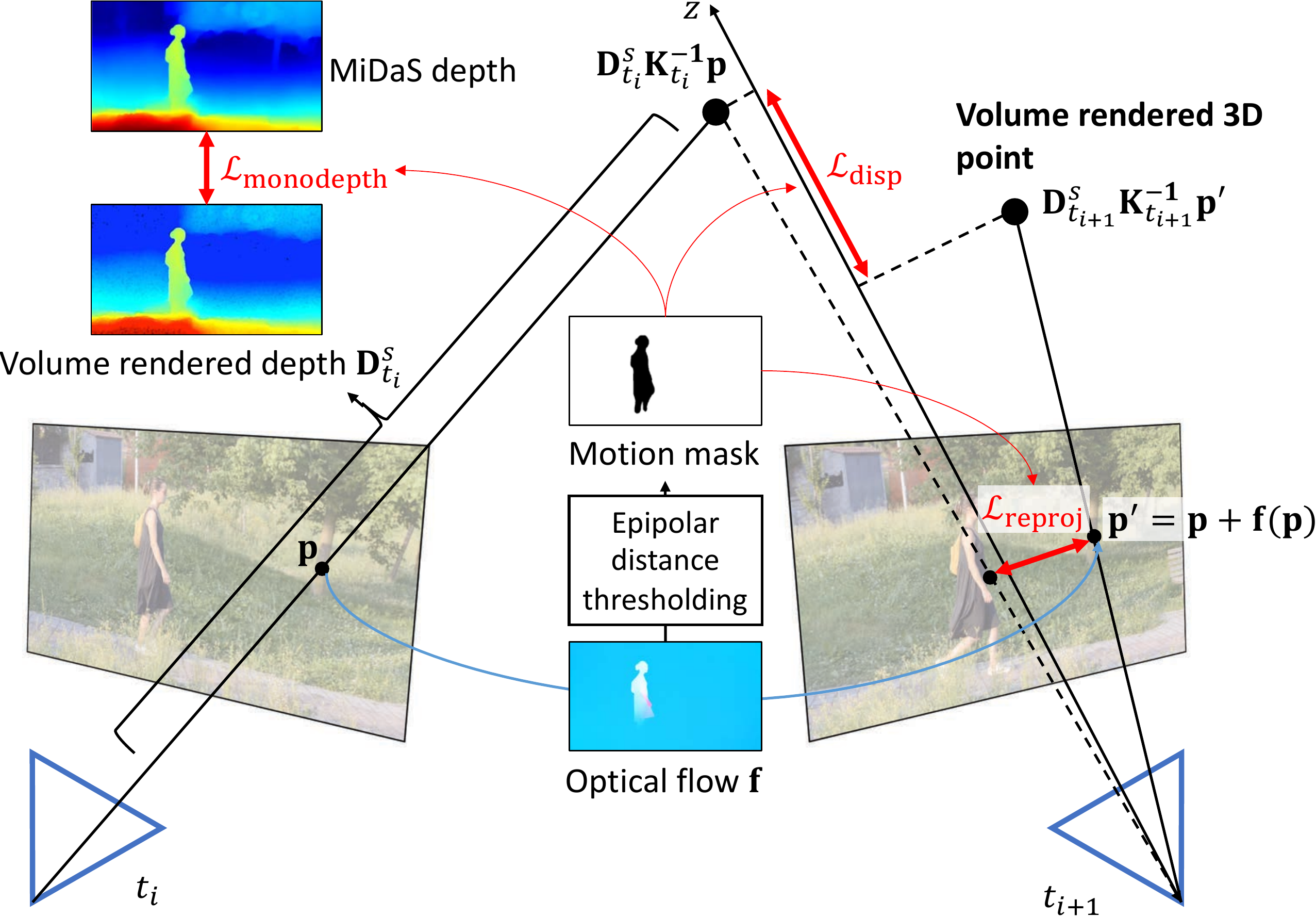} & \includegraphics[height=0.3\textwidth]{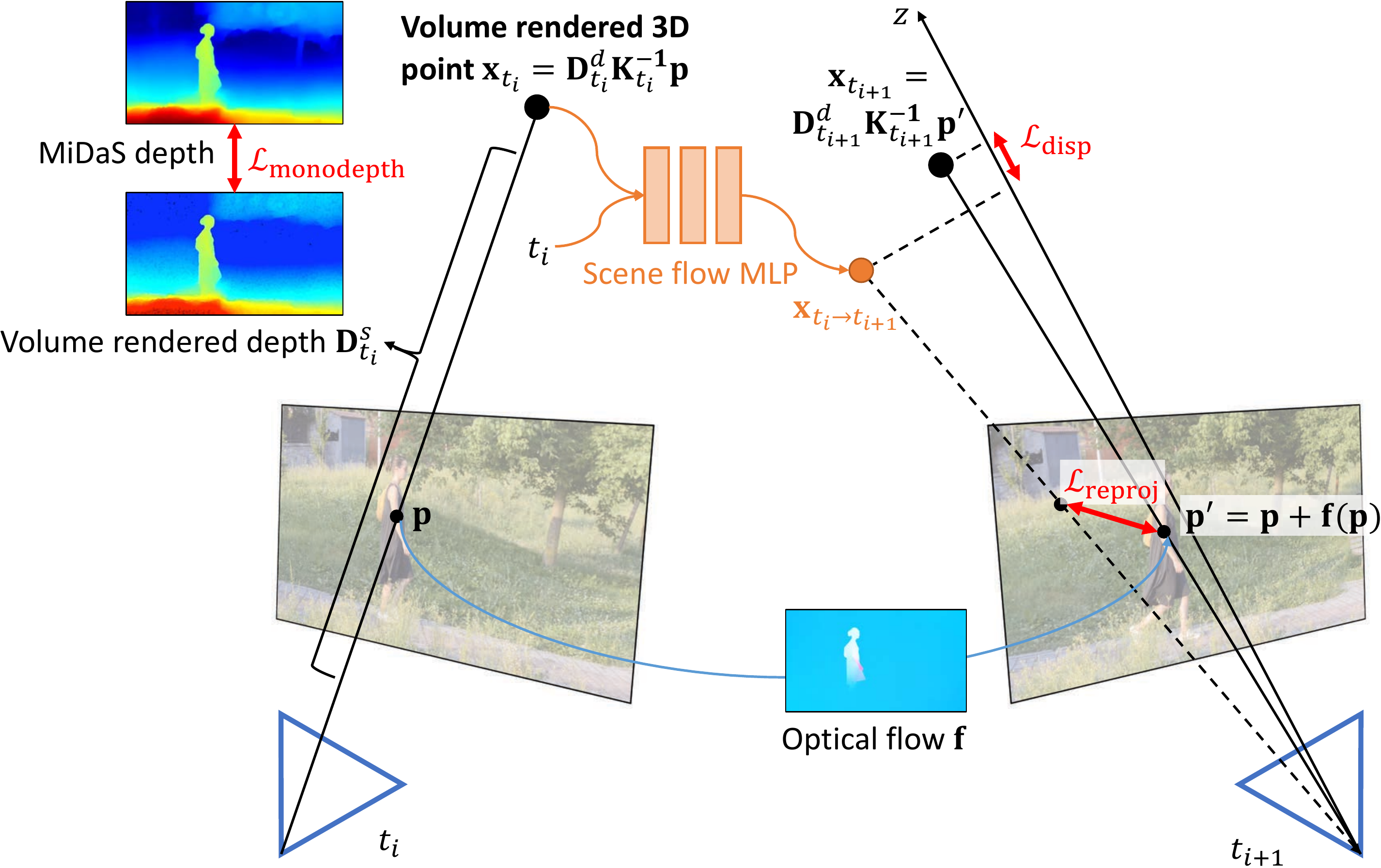} \\
    (a) Static radiance field reconstruction and pose estimation & (b) Dynamic radiance field reconstruction
    \end{tabular}
    }
\vspace{-2mm}
    \caption{\textbf{Training losses.} For both the (a) static and (b) dynamic parts, we introduce three auxiliary losses to encourage the consistency of the modeling: reprojection loss, disparity loss, and monocular depth loss. The reprojection loss encourages the projection of the 3D volume rendered points onto neighbor frames to be similar to the pre-calculated flow. The disparity loss forces the volume rendered 3D points from two corresponding points of neighbor frames to have similar z values. Finally, the monocular depth loss calculates the scale- and shift-invariant loss between the volume rendered depth and the pre-calculated MiDaS depth. (a) We use the motion mask to exclude the dynamic regions from the loss calculation. (b) We use a scene flow MLP to model the 3D movement of the volume rendered 3D points.
}
\label{fig:losses}
\end{figure*}

\section{Method}
\label{sec:method}
In this section, we first briefly introduce the background of neural radiance fields and their extension of camera pose estimation and dynamic scene representation in~\secref{preliminaries}. 
We then describe the overview of our method in~\secref{overview}. 
Next, we discuss the details of camera pose estimation with the static radiance field reconstruction in~\secref{poses}. 
After that, we show how to model the dynamic scene in~\secref{dynamic}. 
Finally, we outline the implementation details in~\secref{details}.

\subsection{Preliminaries}
\label{sec:preliminaries}
\topic{NeRF.} Neural radiance fields (NeRF)~\cite{mildenhall2020nerf} represent a static 3D scene with implicit MLPs parameterized by $\Theta$ and map the 3D position $(x, y, z)$ and viewing direction $(\theta, \phi)$ to its corresponding color $\mathbf{c}$ and density $\sigma$:
\begin{equation}
(\mathbf{c}, \sigma) = \text{MLP}_{\Theta}(x, y, z, \theta, \phi).
\end{equation}
We can compute the pixel color by applying volume rendering~\cite{kajiya1984ray,drebin1988volume} along the ray $\mathbf{r}$ emitted from the camera origin:
\begin{equation}
\begin{gathered}
\hat{\mathbf{C}}(\mathbf{r}) = \sum_{i=1}^{N} T(i)(1-\text{exp}(-\sigma(i)\delta(i)))\mathbf{c}(i),\\
T(i) = \text{exp}(-\sum_{i-1}^{j}\sigma(j)\delta(j)),
\end{gathered}
\end{equation}
where $\delta(i)$ represents the distance between two consecutive sample points along the ray, $N$ is the number of samples along each ray, and $T(i)$ indicates the accumulated transparency. 
As the volume rendering procedure is differentiable, we can optimize the radiance fields by minimizing the reconstruction error between the rendered color $\hat{\mathbf{C}}$ and the ground truth color $\mathbf{C}$:
\begin{equation}
\mathcal{L} = \left \| \hat{\mathbf{C}}(\mathbf{r}) - \mathbf{C}(\mathbf{r}) \right \|_{2}^{2}.
\end{equation}

\topic{Explicit neural voxel radiance fields.} 
Although with compelling rendering quality, NeRF-based methods model the scene with implicit representations such as MLPs for high storage efficiency. 
These methods, however, are very slow to train. 
To overcome this drawback, recent methods~\cite{sun2022direct,fridovich2022plenoxels,muller2022instant,chen2022tensorf} propose to model the radiance fields with explicit voxels. 
Specifically, these methods replace the mapping function with voxel grids and directly optimize the features sampled from the voxels. 
They usually apply shallow MLPs to handle the view-dependent effects. 
By eliminating the heavy usage of the MLPs, the training time of these methods reduces from days to hours. 
We also leverage explicit representation in this work.

\subsection{Method Overview}
\label{sec:overview}
We show our proposed framework in~\figref{overview}. 
Given an input video sequence with $N$ frames, our method jointly optimizes the camera poses, focal length, and static and dynamic radiance fields. 
We represent both the static and dynamic parts with explicit neural voxels $\mathbf{V}^s$ and $\mathbf{V}^d$, respectively. 
The static radiance fields are responsible for reconstructing the static scene and estimating the camera poses and focal length. 
At the same time, the goal of dynamic radiance fields is to model the scene dynamics in the video (usually caused by moving objects).


\subsection{Camera Pose Estimation}
\label{sec:poses}
\topic{Motion mask generation.}
Excluduing dynamic regions in the video helps improve the robustness of camera pose estimation.
Existing methods~\cite{li2021neural} often leverage off-the-shelf instance segmentation methods such as Mask R-CNN~\cite{he2017mask} to mask out the common moving objects. 
However, many moving objects are hard to detect/segment in the input video, such as drifting water or swaying trees.
Therefore, in addition to the masks from Mask R-CNN, we also estimate the fundamental matrix using the optical flow from consecutive frames. 
We then calculate and threshold the Sampson distance (the distance of each pixel to the estimated epipolar line) to obtain a binary motion mask.
Finally, we combine the results from Mask R-CNN and epipolar distance thresholding to obtain our final motion masks.

\begin{figure}[]
\centering
\centering
\resizebox{1.0\columnwidth}{!}
{
\begin{tabular}{cc}
\includegraphics[height=0.3\columnwidth]{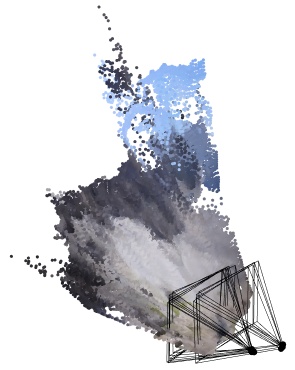} & \includegraphics[height=0.3\columnwidth]{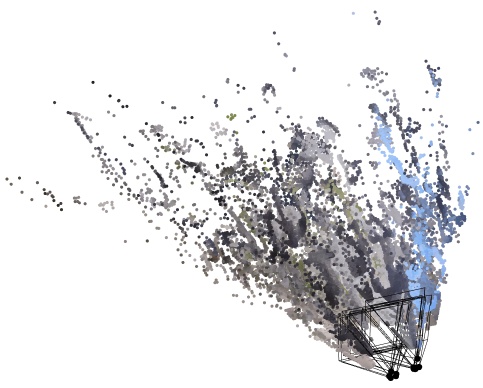} \\
(a) w/o coarse-to-fine & (b) w/o monocular depth prior \\
\includegraphics[height=0.3\columnwidth]{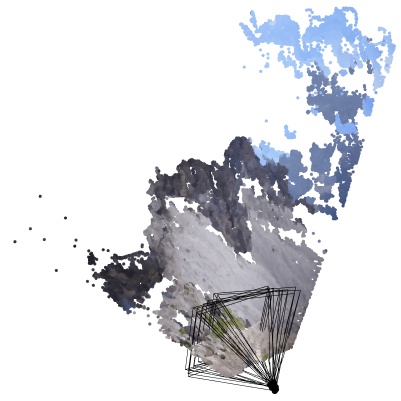} & \includegraphics[height=0.3\columnwidth]{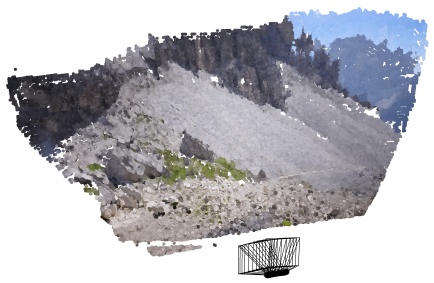} \\
(c) w/o late viewing direction conditioning & (d) Full model \\
\end{tabular}
}
\vspace{-2mm}
\caption{\textbf{The impact of design choices on camera pose estimation.} (a) No coarse-to-fine strategy leads to sub-optimal solutions. (b) No single-image depth prior results in poor scene geometry for challenging camera trajectories. (c) The absence of late viewing direction conditioning leads to wrong geometry and poses due to minimizing photometric loss instead of consistent voxel space using MLP. (d) Our proposed method incorporates all components and yields reasonable scene geometry and camera trajectory.}
\label{fig:pose_ablation}
\end{figure}

\topic{Coarse-to-fine static scene reconstruction.} 
The first part of our method is reconstructing the static radiance fields along with the camera poses. 
We jointly optimize the 6D camera poses $[R|t]_i, i \in [1..N]$ and the focal length $f$ shared by all input frames simultaneously. 
Similar to existing pose estimation methods~\cite{lin2021barf}, we optimize the static scene representation in a coarse-to-fine manner. 
Specifically, we start with a smaller static voxel resolution and progressively increase the voxel resolution during the training. 
This coarse-to-fine strategy is essential to the camera pose estimation as the energy surface will become smoother. 
Thus, the optimizer will have less chance of getting stuck in sub-optimal solutions (\figref{pose_ablation}(a) vs. \figref{pose_ablation}(d)).

\topic{Late viewing direction conditioning.}
As our primary supervision is the photometric consistency loss, the optimization could bypass the neural voxel and directly learn a mapping function from the viewing direction to the output sample color. 
Therefore, we choose to fuse the viewing direction only in the last layer of the color MLP as shown in~\figref{overview}. 
This design choice is critical because we are reconstructing not only the scene geometry but also the camera poses.
\figref{pose_ablation}(c) shows that without the late viewing direction conditioning, the optimization could minimize the photometric loss by optimizing the MLP and lead to erroneous camera poses and geometry estimation.



\topic{Losses.} 
We minimize the photometric loss between the prediction $\hat{\mathbf{C}}^s(\mathbf{r})$ and the captured images in the static regions:
\begin{equation}
    \mathcal{L}^s_c = \left \| (\hat{\mathbf{C}}^s(\mathbf{r}) - \mathbf{C}(\mathbf{r})) \cdot (1-\mathbf{M}(\mathbf{r})) \right \|_2^2,
\end{equation}
where $\mathbf{M}$ denotes the motion mask.

To handle casually-captured but challenging camera trajectories such as fast-moving or pure rotating, we introduce auxiliary losses to regularize the training, similar to \cite{gao2021dynamic,li2021neural}.

(1) Reprojection loss $\mathcal{L}_\text{reproj}^s$: 
We use 2D optical flow estimated by RAFT~\cite{teed2020raft} to guide the training. 
First, we volume render all the sampled 3D points along a ray to generate a \emph{surface} point. 
We then reproject this point onto its neighbor frame and calculate the reprojection error with the correspondence estimated from RAFT.

(2) Disparity loss $\mathcal{L}_\text{disp}^s$: 
Similar to the reprojection loss above, we also regularize the error in the z-direction (in the camera coordinate). 
We volume render the two corresponding points into 3D space and calculate the error of the z component. 
As we care more about the near than the far, we compute this loss in the inverse-depth domain.

(3) Monocular depth loss $\mathcal{L}_\text{monodepth}^s$: 
The two losses above cannot handle pure rotating cameras and often lead to the incorrect camera poses and geometry (\figref{pose_ablation}(b)). 
We enforce the depth order from multiple pixels of the same frame to match the order of a monocular depth map. 
We pre-calculate the depth map using MiDaSv2.1~\cite{ranftl2020towards}. 
The depth prediction from MiDaS is up to an unknown scale and shift. 
Therefore, we use the same scale- and shift-invariant loss in MiDaS to constrain our rendered depth values.

We illustrate these auxiliary losses in~\figref{losses}(a). 
Since the optical flow and depth map may not be accurate, we apply annealing for the weights of these auxiliary losses during the training. 
As the input frames contain dynamic objects, we need to mask out all the dynamic regions while applying all these losses and the L2 reconstruction loss. 
The final loss for the static part is:
\begin{equation}
    \mathcal{L}^s = \mathcal{L}^s_c + \lambda_\text{reproj}^s \mathcal{L}_\text{reproj}^s + \lambda_\text{disp}^s \mathcal{L}_\text{disp}^s + \lambda_\text{monodepth}^s \mathcal{L}_\text{monodepth}^s.
\end{equation}


\subsection{Dynamic Radiance Field Reconstruction}
\label{sec:dynamic}
\topic{Handling temporal information.}
To query the time-varying features from the voxel, we first pass the 3D coordinates $(x, y, z)$ along with time index $t_i$ to a coordinate deformation MLP. 
The coordinate deformation MLP predicts the 3D time-varying deformation vectors $(\Delta x, \Delta y, \Delta z)$. 
We then add these deformations onto the original coordinates to get the \emph{deformed} coordinates $(x', y', z')$. 
This deformation MLP indicates that the voxel is a canonical space and that each corresponding 3D point from a different time should point to the same position in this voxel space. 
We design the deformation MLP to deform the 3D points from the original camera space to the canonical voxel space.

However, using a single compact canonical voxel to represent the entire sequence along the temporal dimension is very challenging. 
Therefore, we further introduce~\emph{time-dependent MLPs} to enhance the queried features from the voxel to predict time-varying color and density. 
Note that the time-dependent MLPs with only two to three layers are much shallower than the ones in other dynamic view synthesis methods~\cite{li2021neural,gao2021dynamic} as the purpose of the MLPs here is further to enhance the queried features from the canonical voxel. 
Most of the time-varying effects are still carried out by the coordination deformation MLP. 
We show the above architecture at the bottom of~\figref{overview}. And the photometric training loss for the dynamic part is:
\begin{equation}
    \mathcal{L}^d_c = \left \| \hat{\mathbf{C}}^d(\mathbf{r}) - \mathbf{C}(\mathbf{r}) \right \|_2^2,
\end{equation}

\begin{figure}[]
\centering
\scriptsize
\resizebox{\columnwidth}{!}{%
\begin{tabular}{ccccc}
\makecell[b]{\includegraphics[width=0.2\columnwidth]{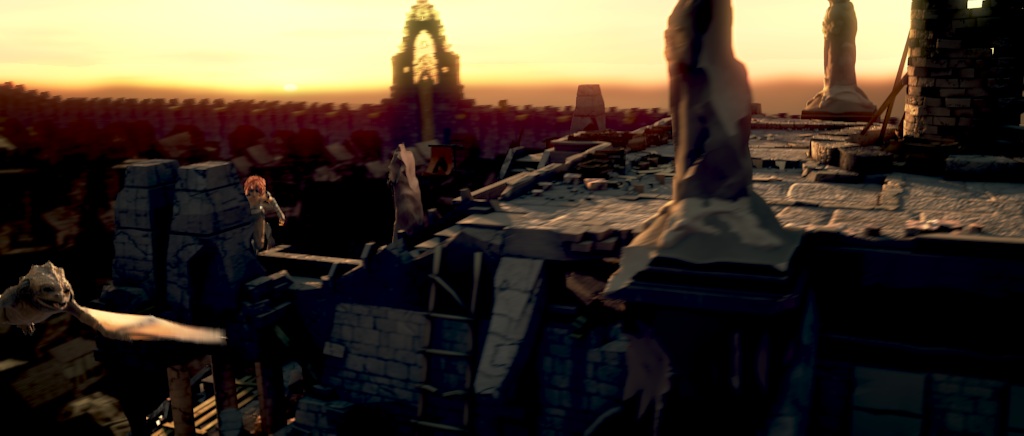} \\ \vspace{2mm} \includegraphics[width=0.2\columnwidth]{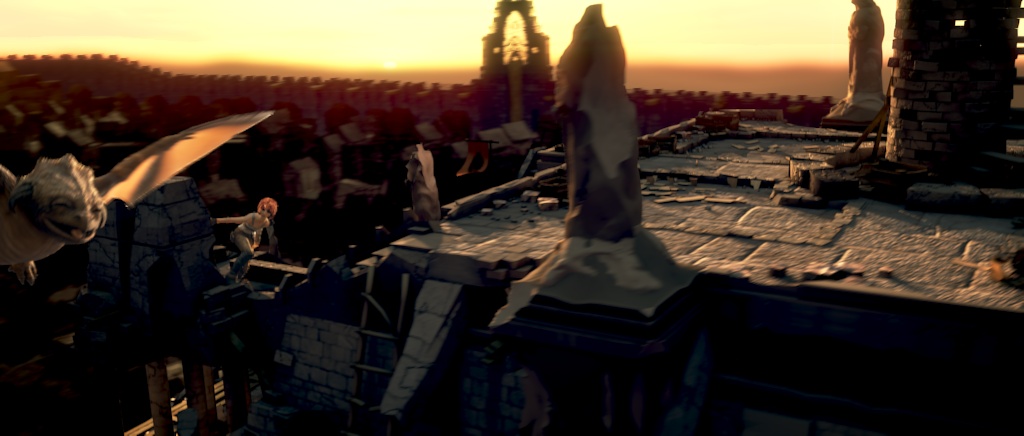}} & \includegraphics[width=0.2\columnwidth]{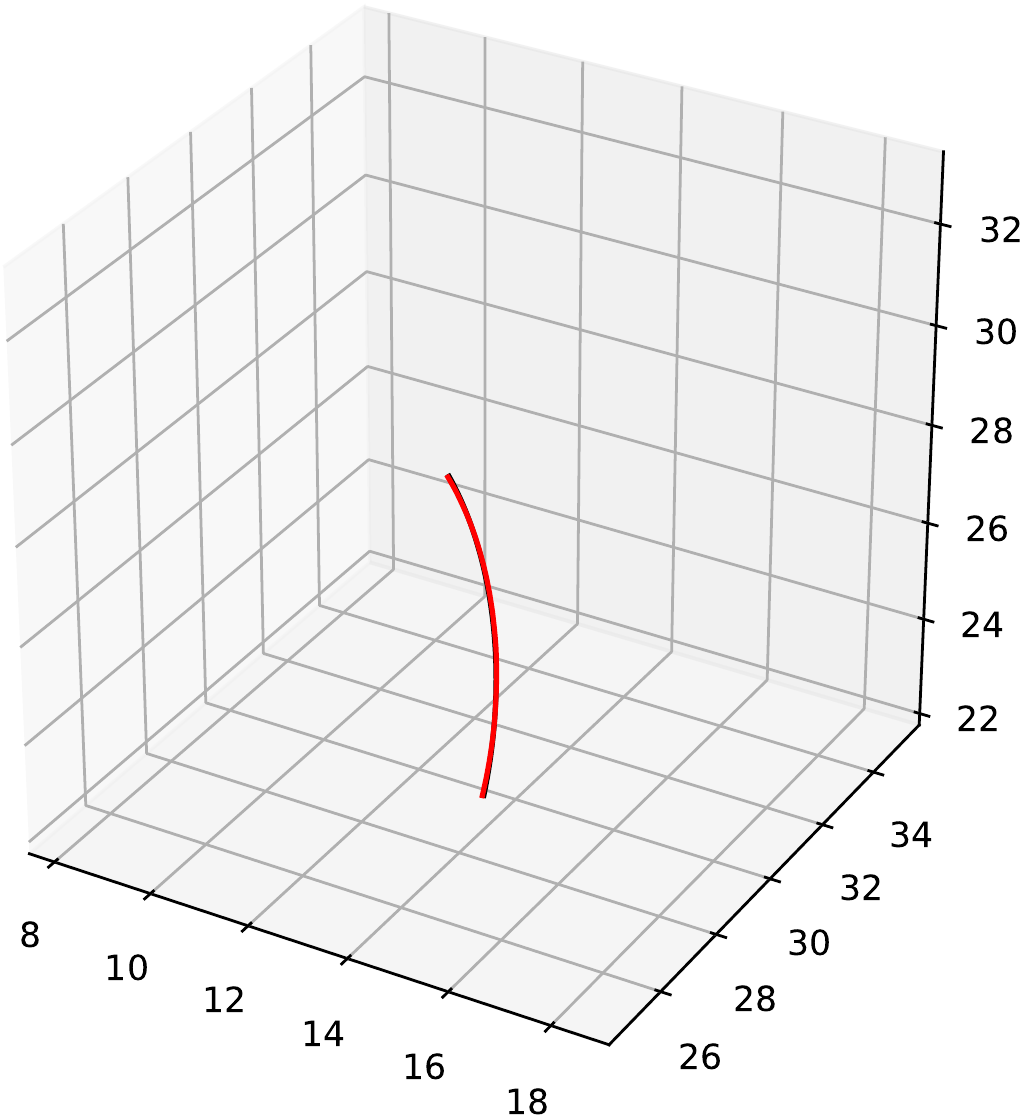} &
\includegraphics[width=0.2\columnwidth]{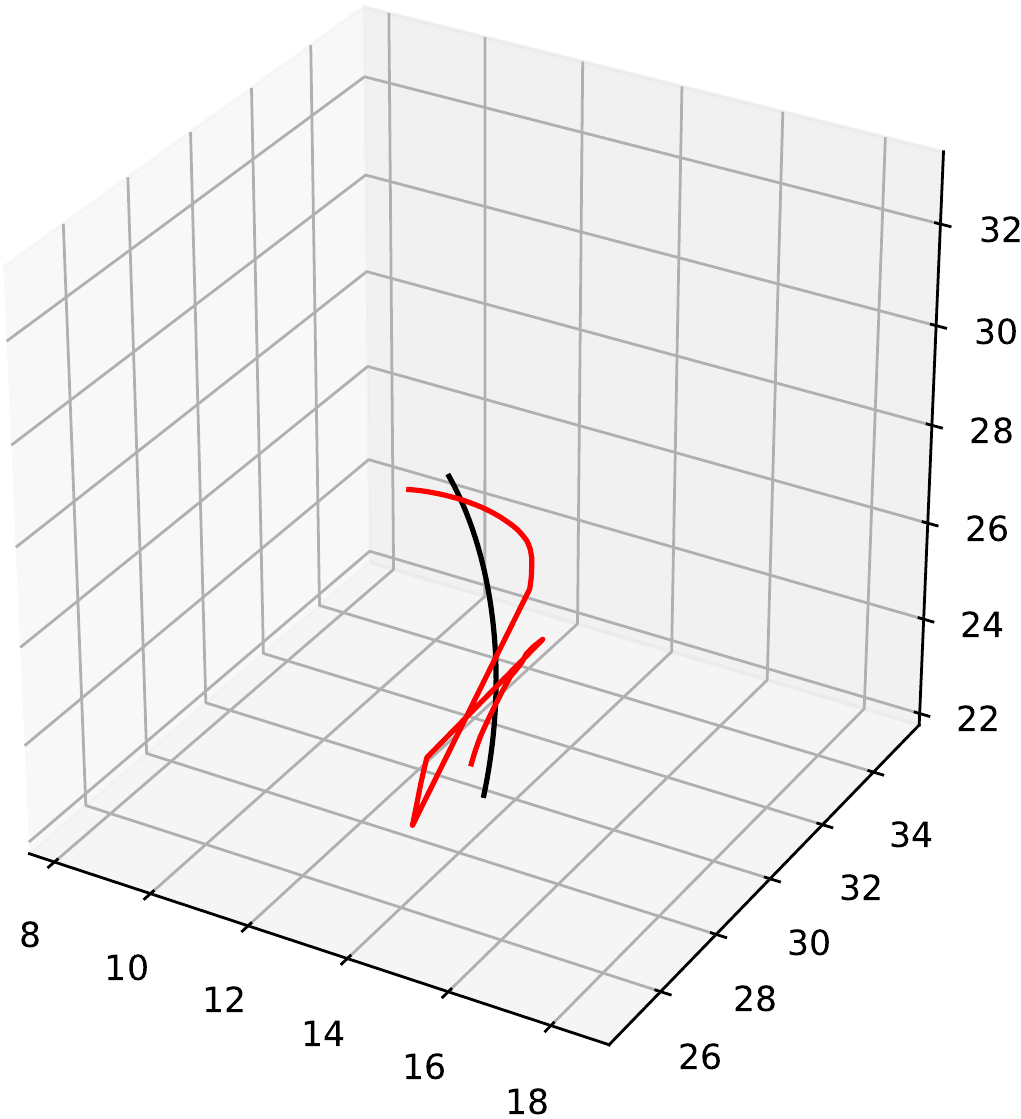} &
\includegraphics[width=0.2\columnwidth]{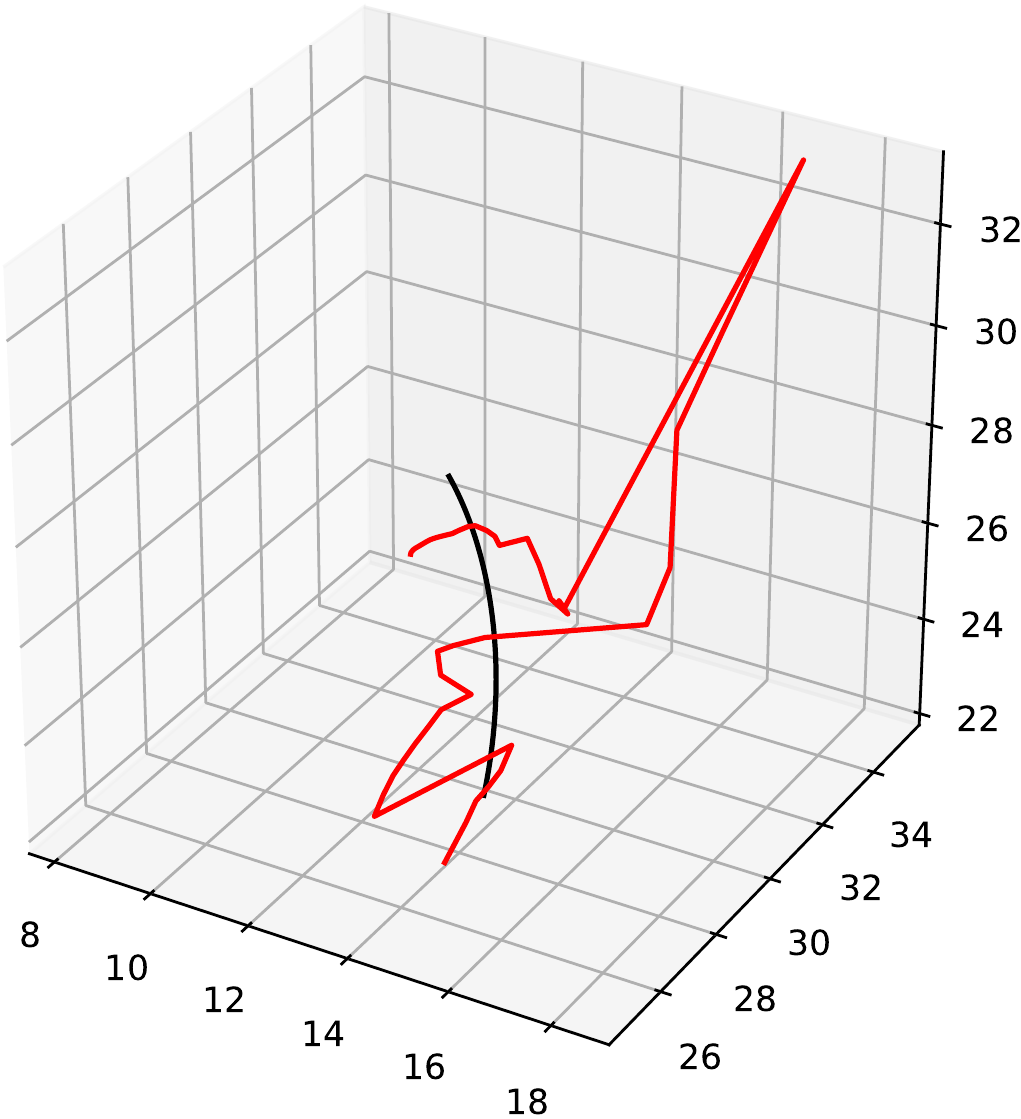} & 
\includegraphics[width=0.2\columnwidth]{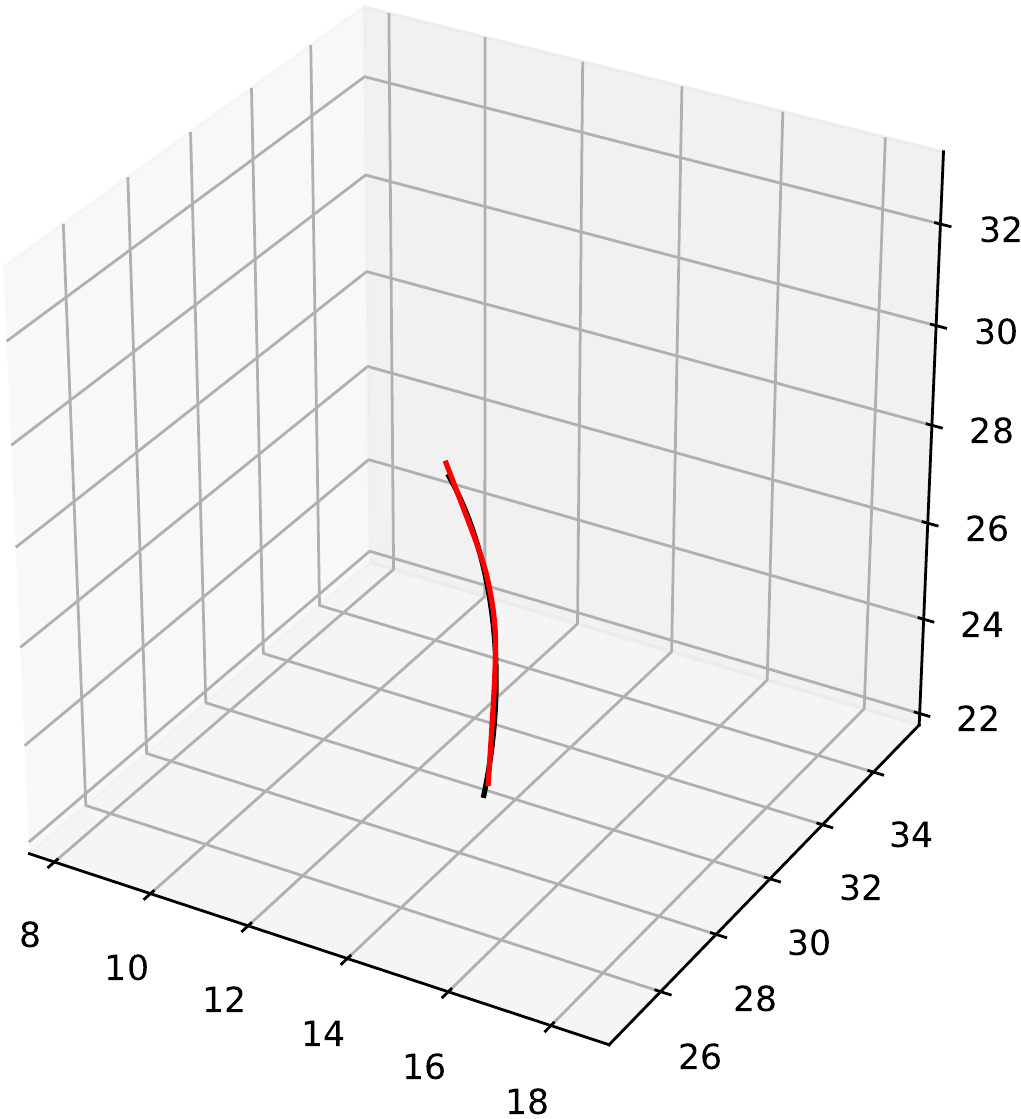} \\
\makecell[b]{\includegraphics[width=0.2\columnwidth]{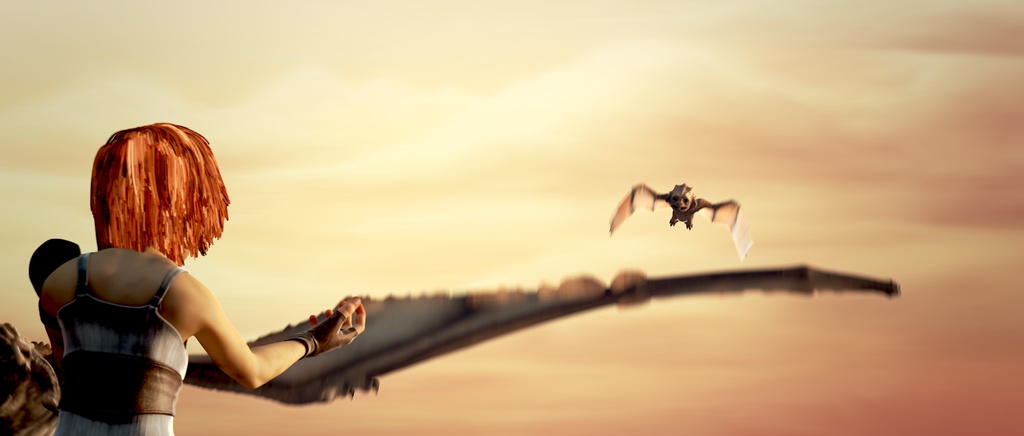} \\ \vspace{2mm} \includegraphics[width=0.2\columnwidth]{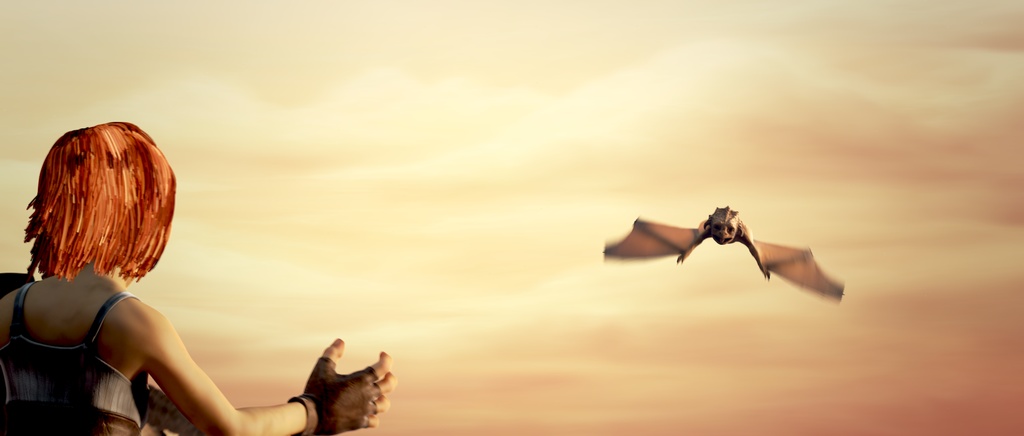}} & \includegraphics[width=0.2\columnwidth]{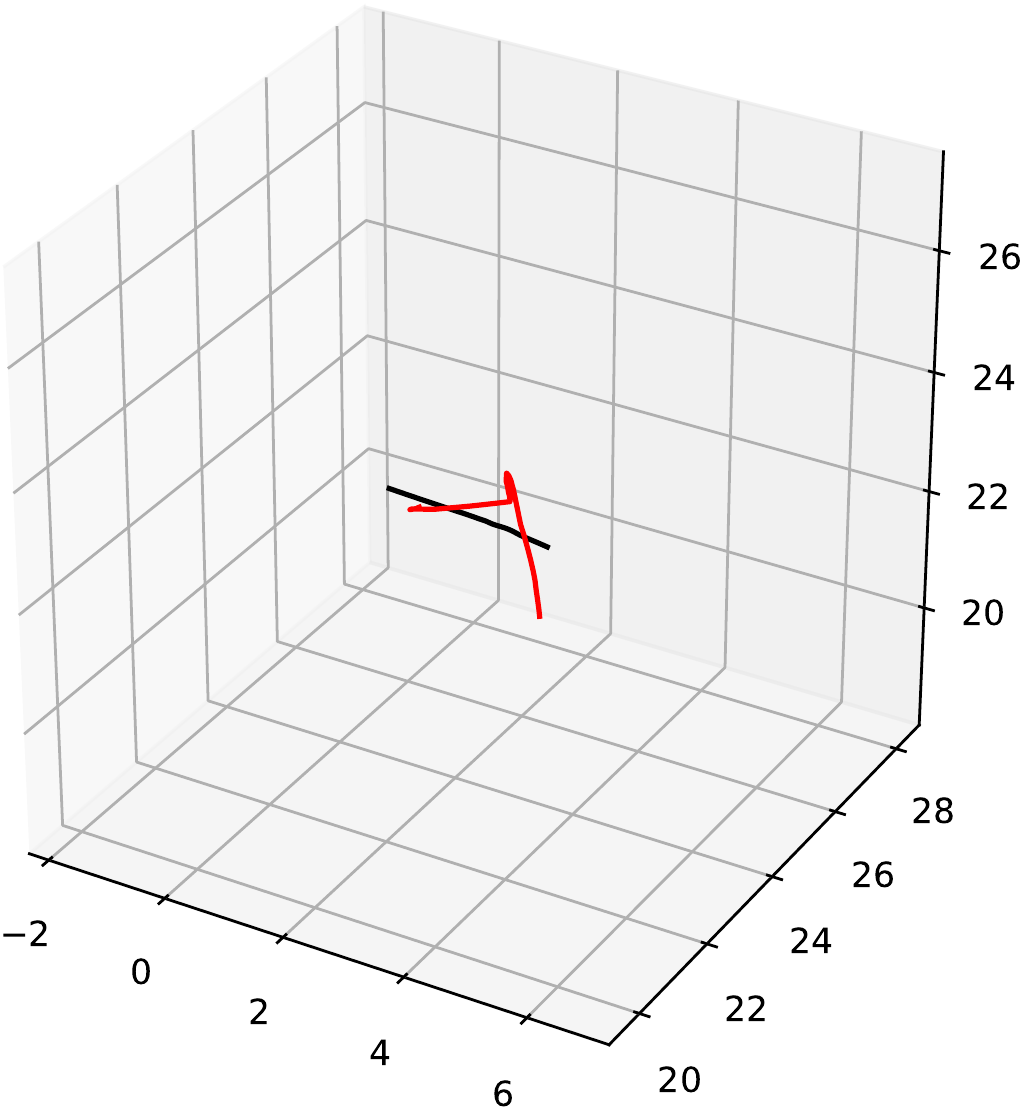} & 
\includegraphics[width=0.2\columnwidth]{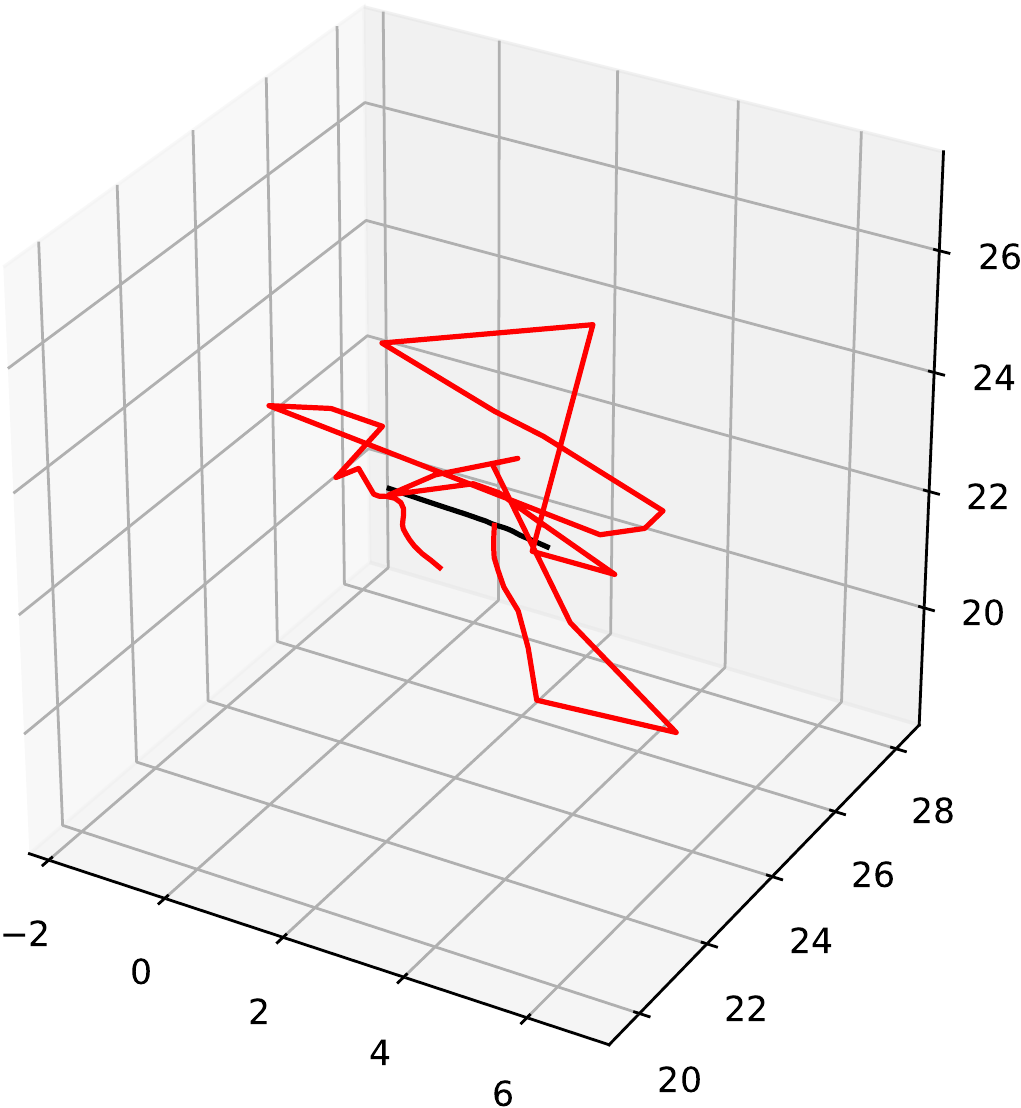} &
\includegraphics[width=0.2\columnwidth]{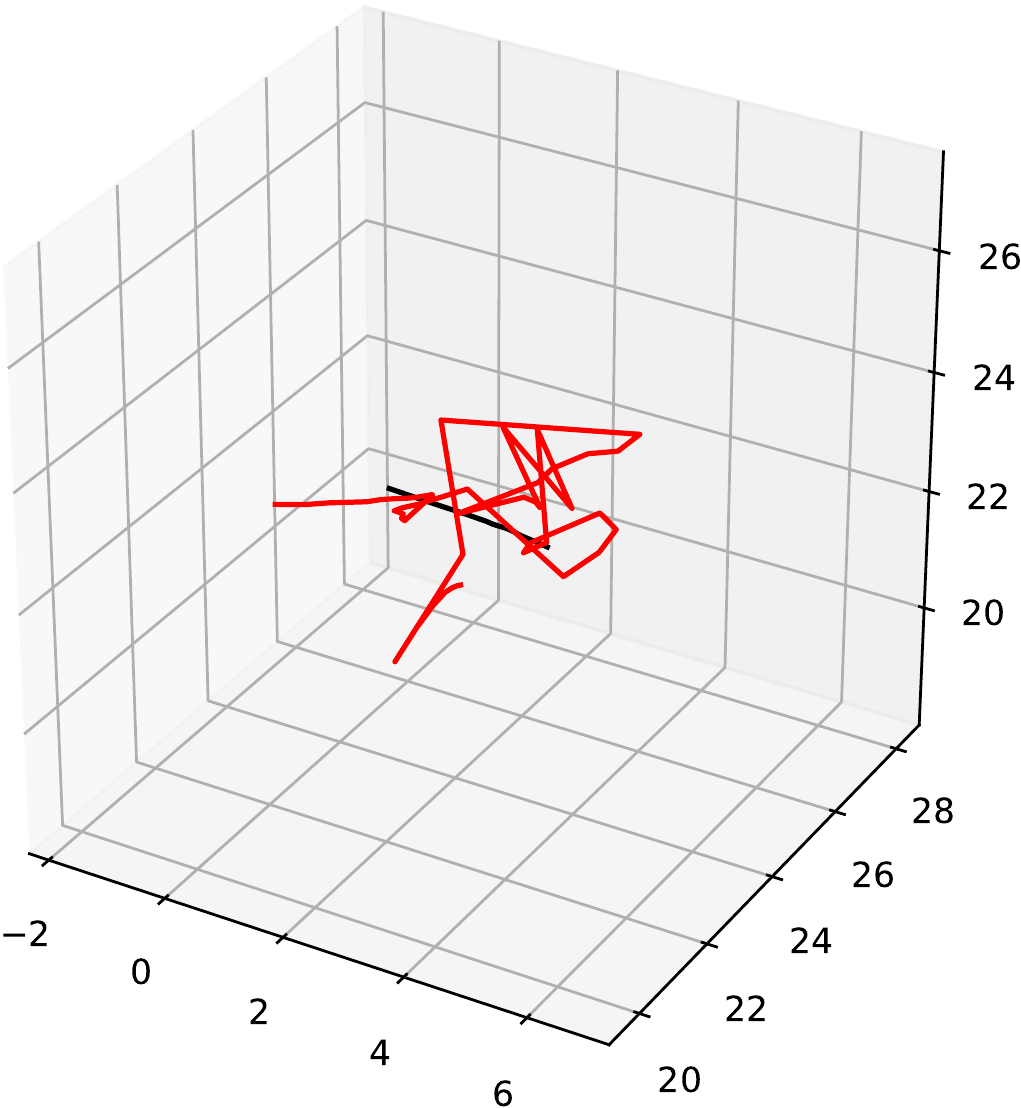} &
\includegraphics[width=0.2\columnwidth]{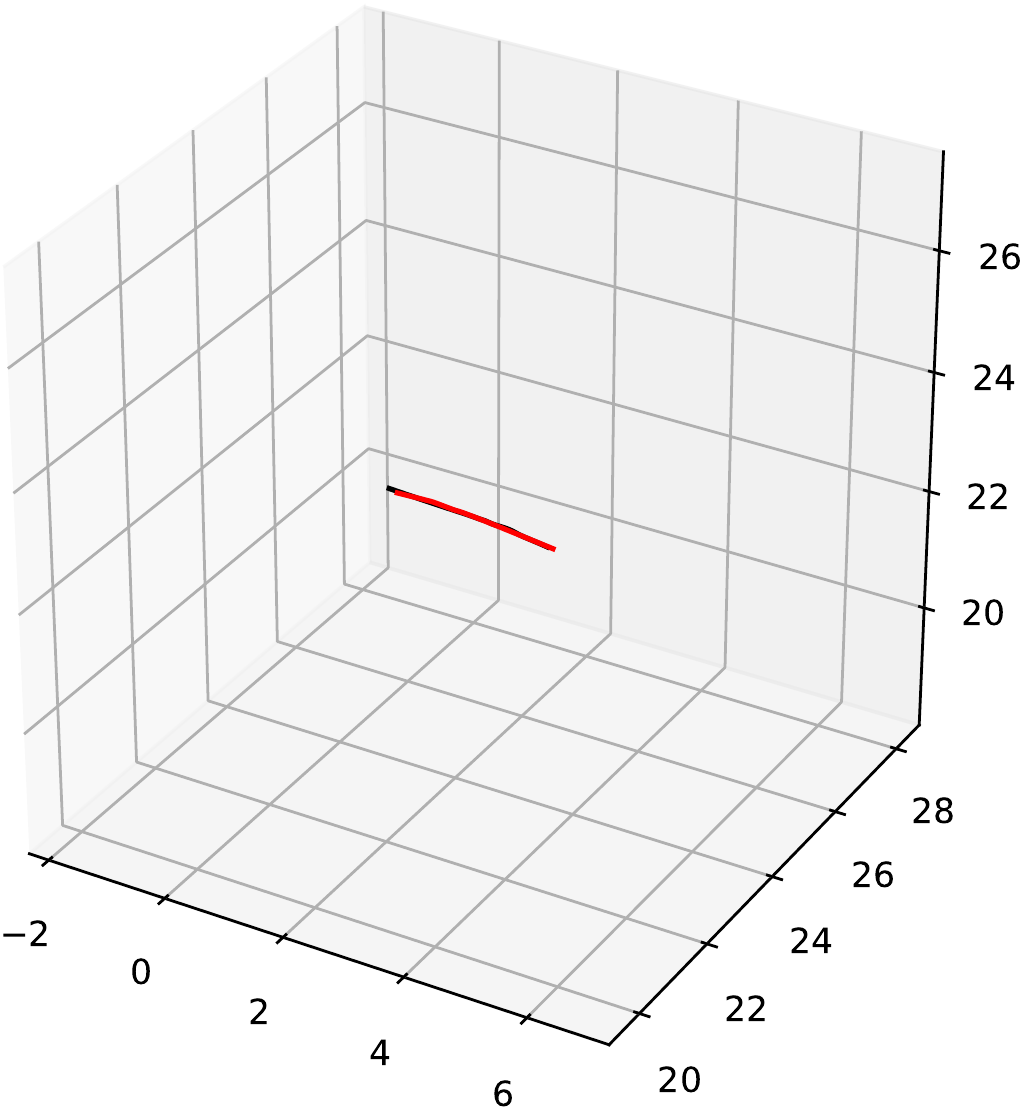} \\
\makecell[b]{\includegraphics[width=0.2\columnwidth]{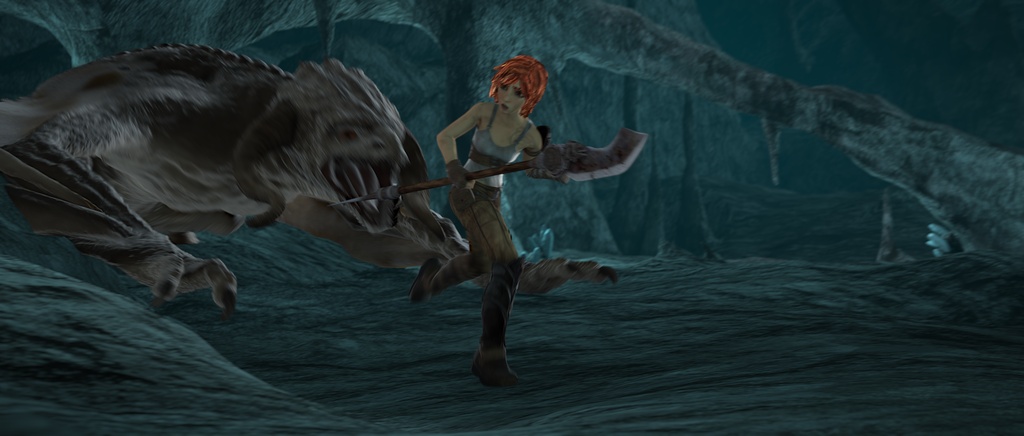} \\ \vspace{2mm} \includegraphics[width=0.2\columnwidth]{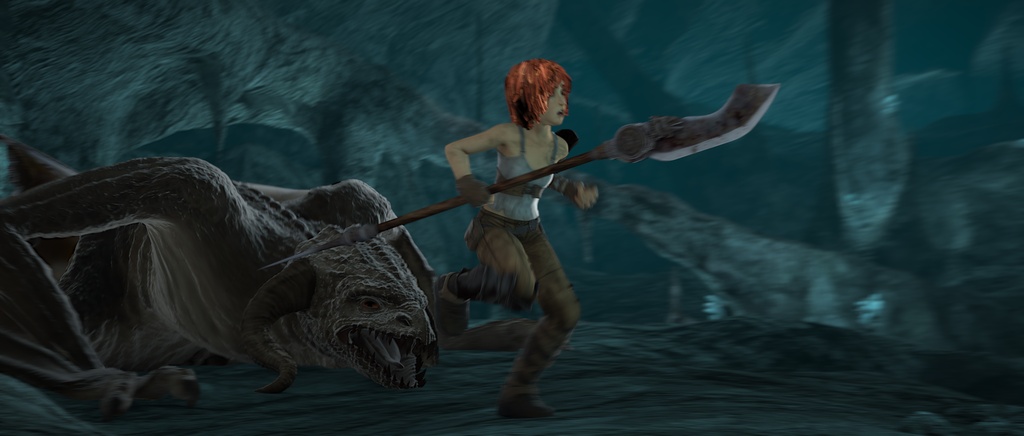}} & \includegraphics[width=0.2\columnwidth]{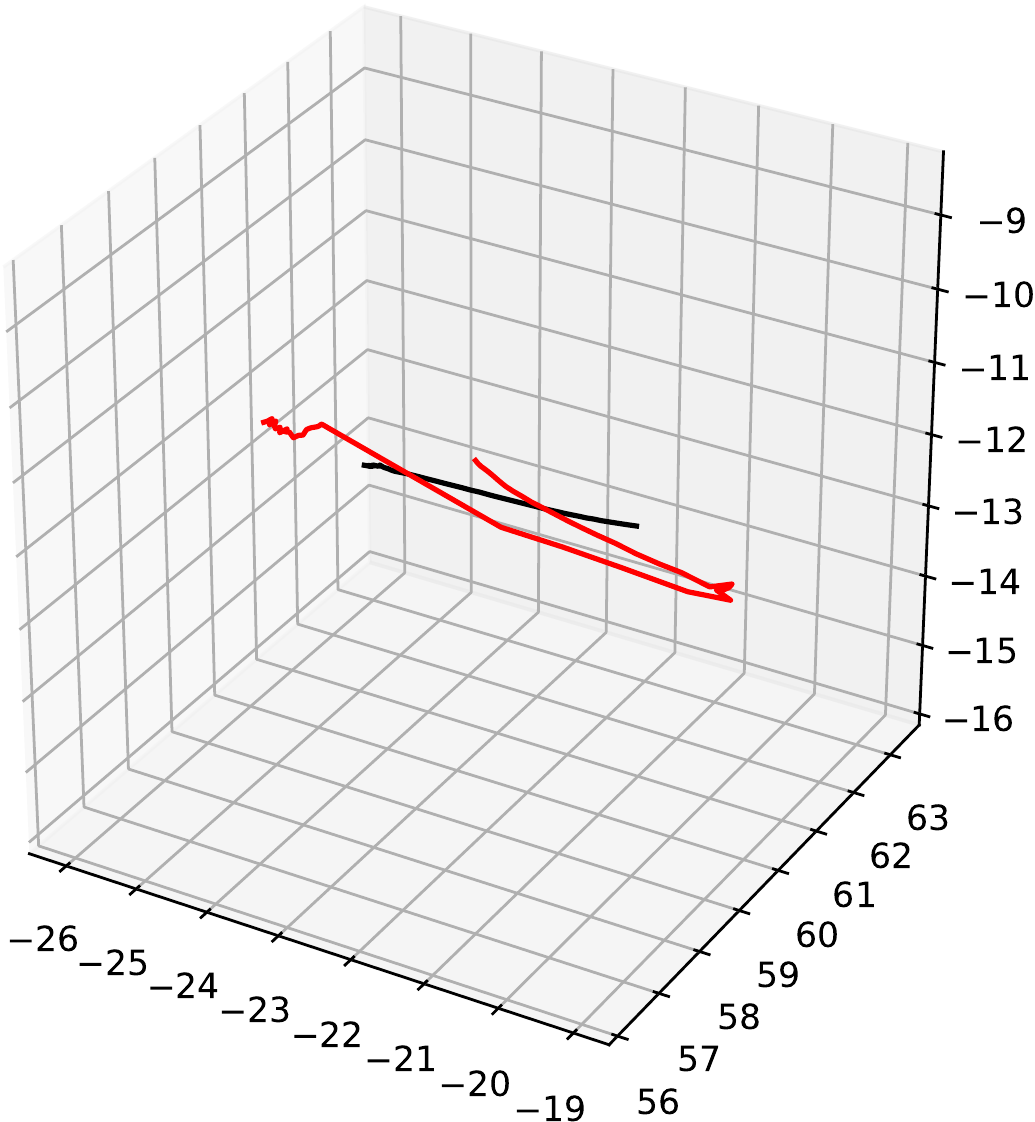} & 
\includegraphics[width=0.2\columnwidth]{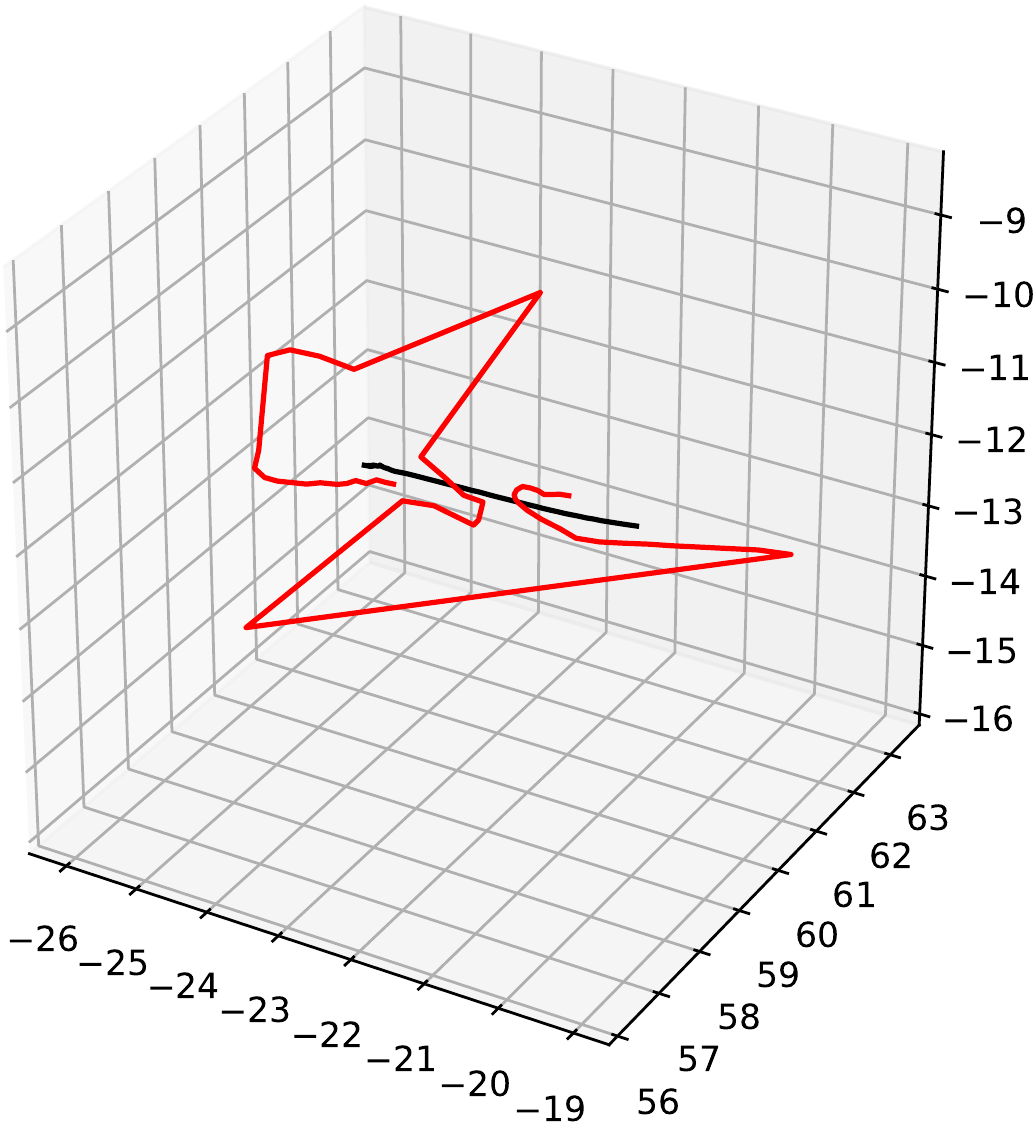} &
\includegraphics[width=0.2\columnwidth]{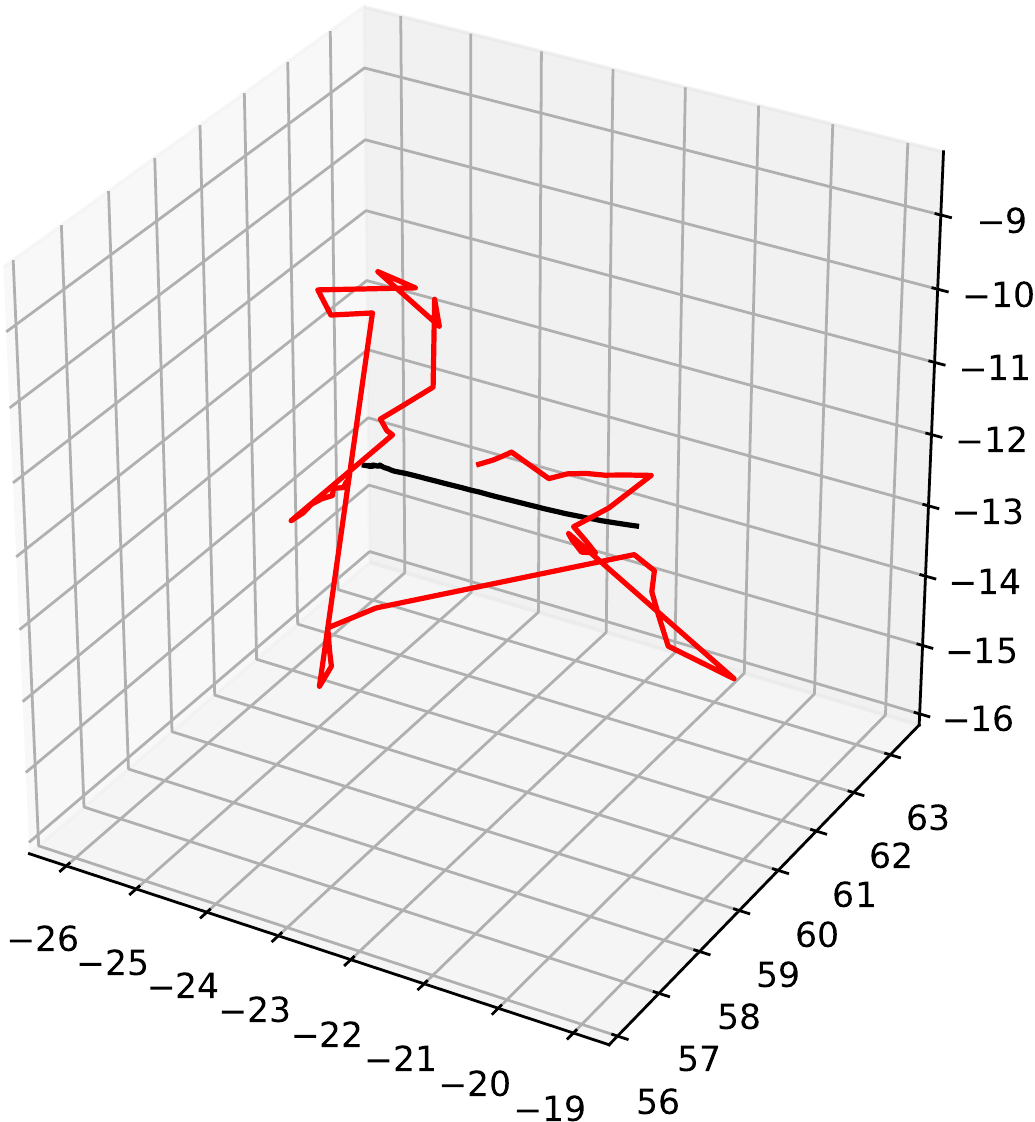} &
\includegraphics[width=0.2\columnwidth]{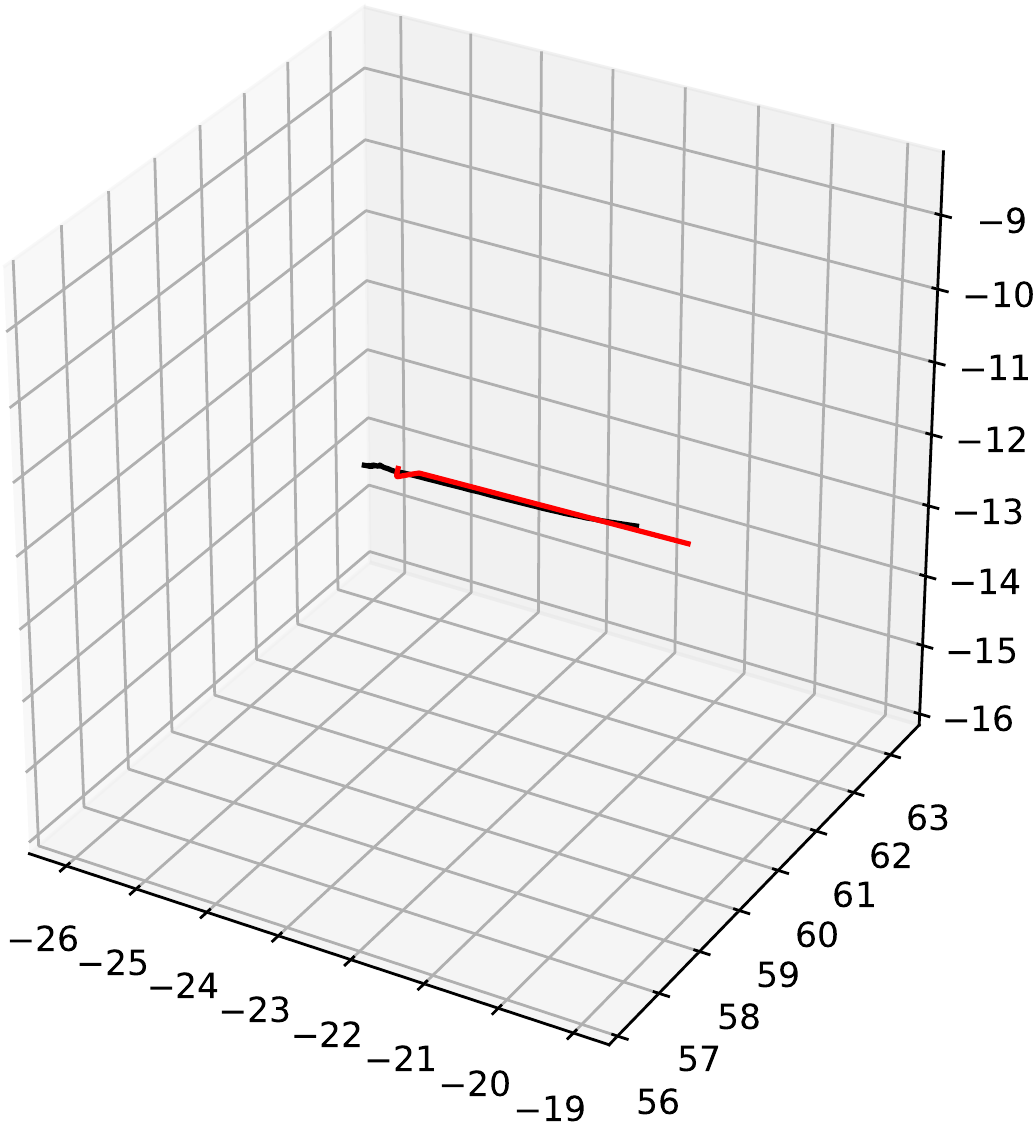} \\
Sample frames & ParticleSfM~\cite{zhao2022particlesfm} & NeRF - -~\cite{wang2021nerf} & BARF~\cite{lin2021barf} & Ours \\
\end{tabular}%

\vspace{-3mm}}
\caption{\textbf{Qualitative results of moving camera localization on the MPI Sintel dataset.}
}
\label{fig:visual_Sintel}
\end{figure}
\begin{table}[]
\centering
\caption{\textbf{Quantitative evaluation of camera poses estimation on the MPI Sintel dataset.} The methods of the top block discard the dynamic components and do not reconstruct the radiance fields; thus they cannot render novel views. We exclude the COLMAP results since it fails to produce poses in 5 out of 14 sequences.
}
\vspace{-2mm}
\label{tab:quantitative_sintel}
\resizebox{\columnwidth}{!}{%
\begin{tabular}{lccc}
\toprule
Method & ATE (m) & RPE trans (m) & RPE rot (deg) \\
\midrule
R-CVD~\cite{kopf2021robust} & 0.360 & 0.154 & 3.443\\
DROID-SLAM~\cite{teed2021droid} & 0.175 & 0.084 & 1.912 \\
ParticleSfM~\cite{zhao2022particlesfm} & \second{0.129} & \first{0.031} & \first{0.535} \\
\midrule
NeRF - -~\cite{wang2021nerf} & 0.433 & 0.220 & 3.088 \\
BARF~\cite{lin2021barf} & 0.447 & 0.203 & 6.353 \\
Ours & \first{0.089} & \second{0.073} & \second{1.313} \\
\bottomrule
\end{tabular}%
}
\end{table}
\begin{figure}[]
\centering
\scriptsize
\resizebox{\columnwidth}{!}{%
\begin{tabular}{ccc}
\includegraphics[width=.32\linewidth]{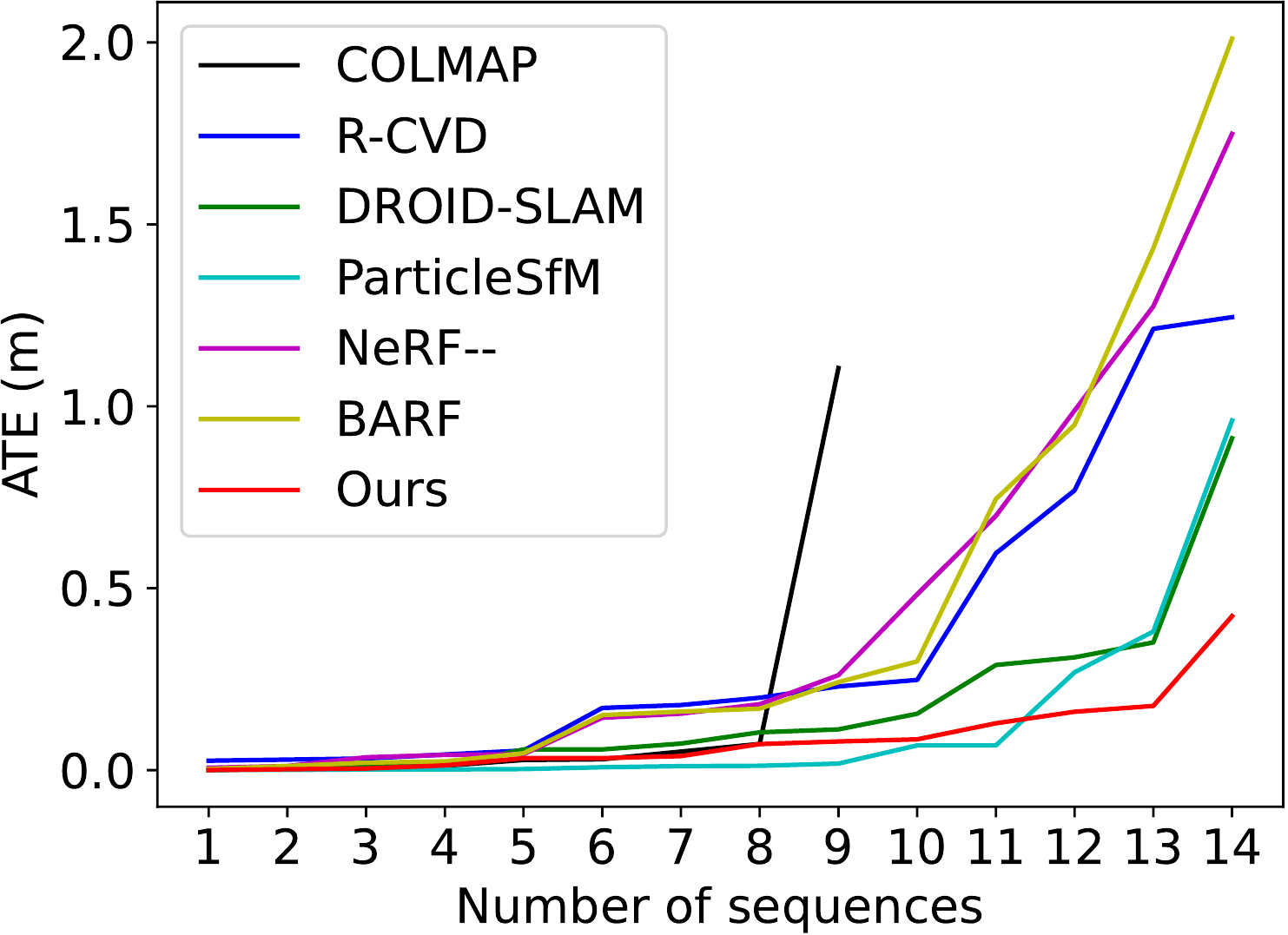} & 
\includegraphics[width=.32\linewidth]{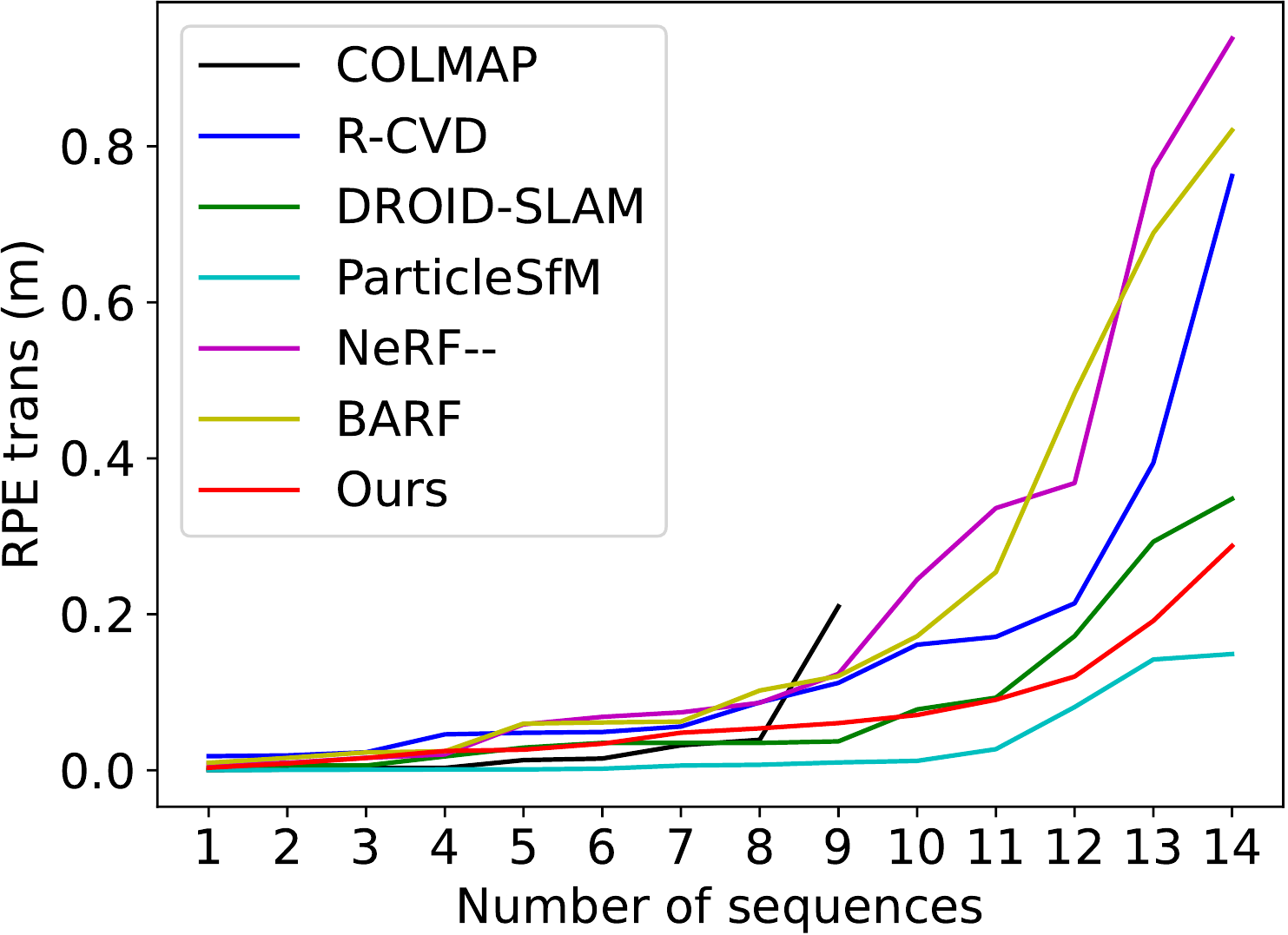} & 
\includegraphics[width=.32\linewidth]{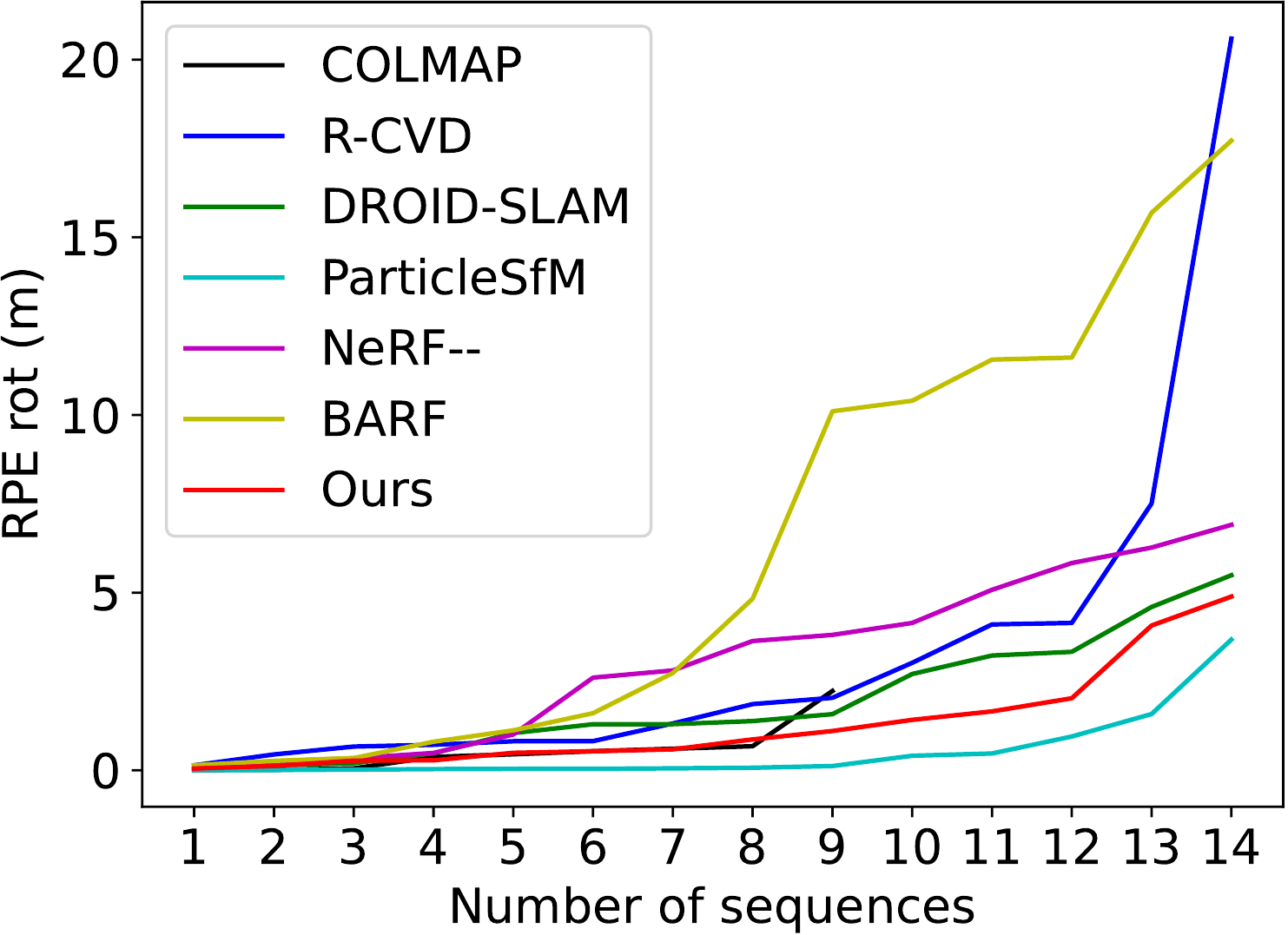} \\
\end{tabular}%
\vspace{-3mm}}
\caption{\textbf{The sorted error plots showing both the \emph{accuracy} and \emph{completeness/robustness} in the MPI Sintel dataset.}
}
\label{fig:curve_Sintel}
\end{figure}


\topic{Scene flow modeling.}
We introduce three losses based on external priors to better model the dynamic movements. 
The three losses are similar to the ones in the static part, but we need to model the movements of the 3D points. 
Therefore, we introduce a scene flow MLP to compensate the 3D motion. 
\begin{equation}
(S_{i \rightarrow i+1}, S_{i \rightarrow i-1}) = \text{MLP}_{\theta_\text{sf}}(x, y, z, t_i),
\end{equation}
where $S_{i \rightarrow i+1}$ represents the 3D scene flow of the 3D point $(x, y, z)$ at time $t_i$.
With the 3D scene flow, we can apply the losses for the dynamic radiance fields. We show the training losses in~\figref{losses}(b). 

(1) Reprojection loss $\mathcal{L}_\text{reproj}^d$: We induce the 2D flow using the poses, depth, and the estimated 3D scene flow from the scene flow MLP. And we compare the error of this induced flow with the one estimated by RAFT.

(2) Disparity loss $\mathcal{L}_\text{disp}^d$: Similar to the disparity loss in the static part, but here we additionally have the 3D scene flow. We get the corresponding points in the 3D space, add the estimated 3D scene flow, and calculate the difference of the z components in the inverse-depth domain.

(3) Monocular depth loss $\mathcal{L}_\text{monodepth}^d$: We calculate scale- and shift-invariant loss between the rendered depth with the pre-calculated depth map using MiDaSv2.1.

We further regularize the 3D motion prediction from the MLP by introducing the smooth and small scene flow loss:
\begin{equation}
\mathcal{L}_{\text{sf}}^{\text{reg}} = \left \| S_{i \rightarrow i+1} + S_{i \rightarrow i-1} \right \|_1 + \left \| S_{i \rightarrow i+1} \right \|_1 + \left \| S_{i \rightarrow i-1} \right \|_1.
\end{equation}
Note that the scene flow MLP is \emph{not} part of the rendering process but part of the losses. 
By representing the 3D scene flow with an MLP and enforcing proper priors, we can make the density prediction better and more reasonable. 
We also detach the gradients from the dynamic radiance fields to the camera poses. 
Finally, we supervise the nonrigidity mask $\mathbf{M}^d$ with motion mask $\mathbf{M}$:
\begin{equation}
\mathcal{L}^d_{m} = \left \| \mathbf{M}^d - \mathbf{M} \right \|_1.
\end{equation}
The overall loss of the dynamic part is:
\begin{equation}
\begin{gathered}
    \mathcal{L}^d = \mathcal{L}^d_c + \lambda_\text{reproj}^d \mathcal{L}_\text{reproj}^d + \lambda_\text{disp}^d \mathcal{L}_\text{disp}^d + \\
    \lambda_\text{monodepth}^d \mathcal{L}_\text{monodepth}^d + \lambda_{\text{sf}}^{\text{reg}} \mathcal{L}_{\text{sf}}^{\text{reg}} + \lambda^d_{m} \mathcal{L}^d_{m}.
\end{gathered}
\end{equation}

\begin{figure}[]
\centering
\footnotesize
\setlength{\tabcolsep}{1pt}
\renewcommand{\arraystretch}{1}
\resizebox{\columnwidth}{!}{%
\begin{tabular}{ccccc}
\raisebox{2.2\normalbaselineskip}[0pt][0pt]{\rotatebox[origin=c]{90}{Ground truth}} &
\includegraphics[width=0.17\textwidth]{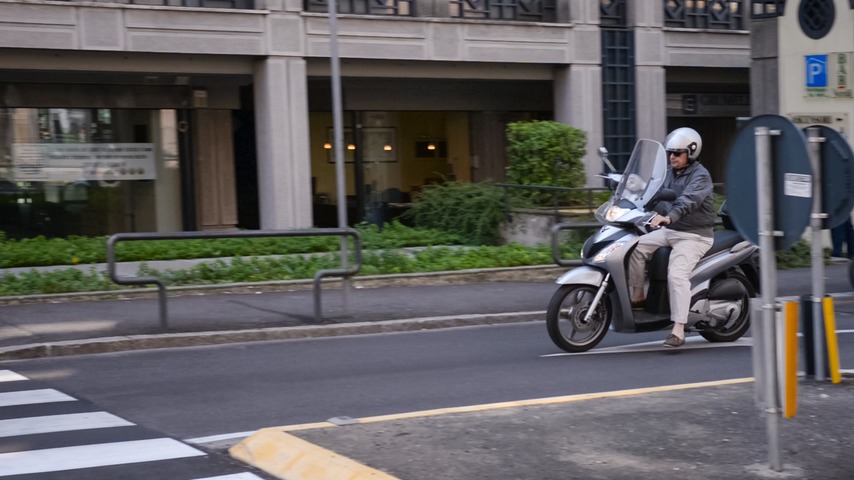} & \includegraphics[width=0.17\textwidth]{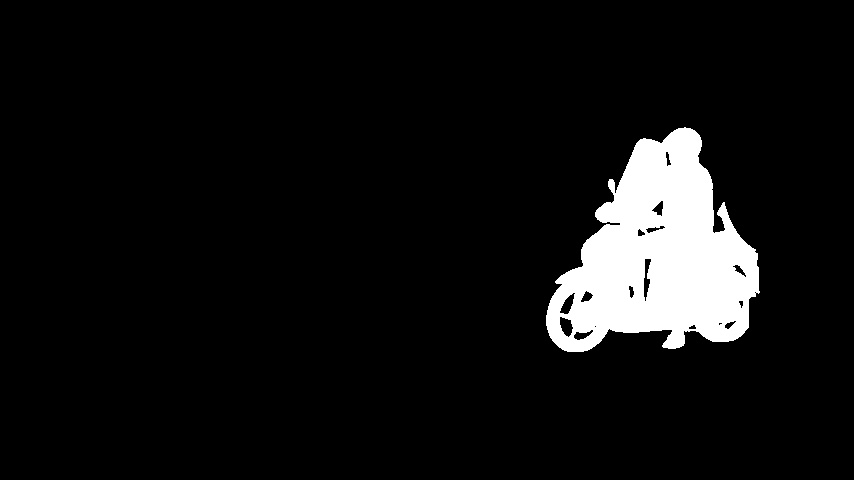} & 
\includegraphics[width=0.17\textwidth]{figures/results/DAVIS_480p/gt/car-roundabout/00014.jpg} & \includegraphics[width=0.17\textwidth]{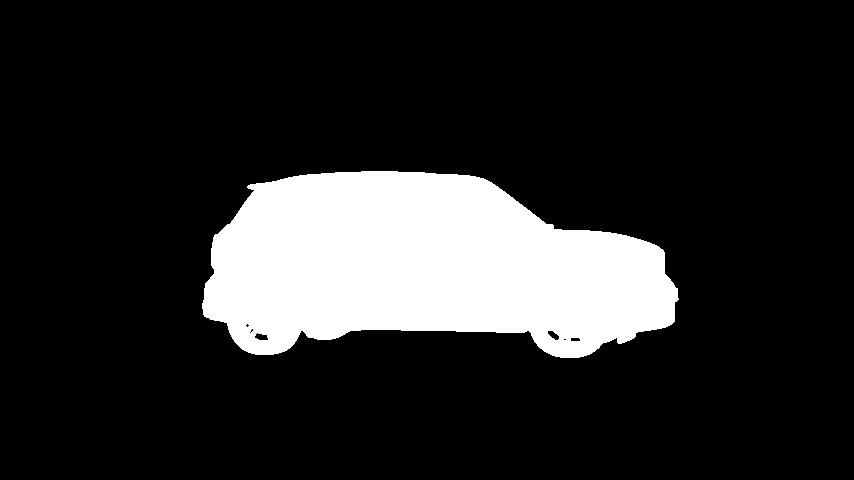} \\
\raisebox{2.2\normalbaselineskip}[0pt][0pt]{\rotatebox[origin=c]{90}{NeRF - -~\cite{wang2021nerf}}} &
\includegraphics[width=0.17\textwidth]{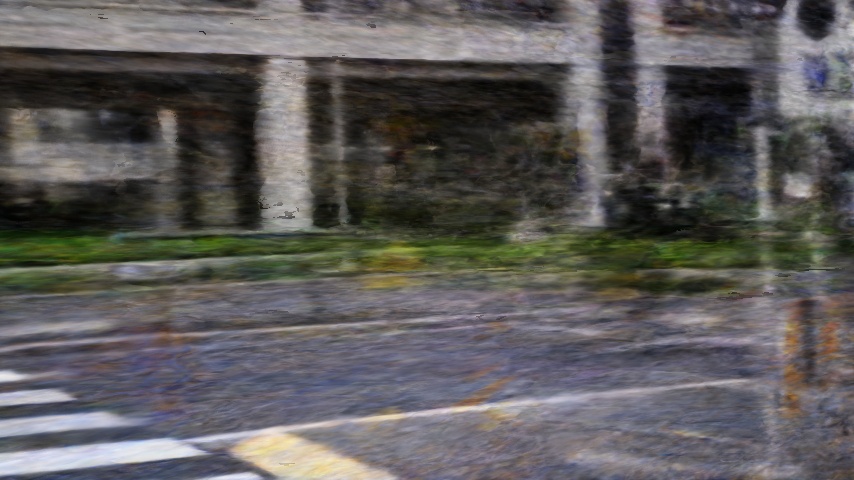} & \includegraphics[width=0.17\textwidth]{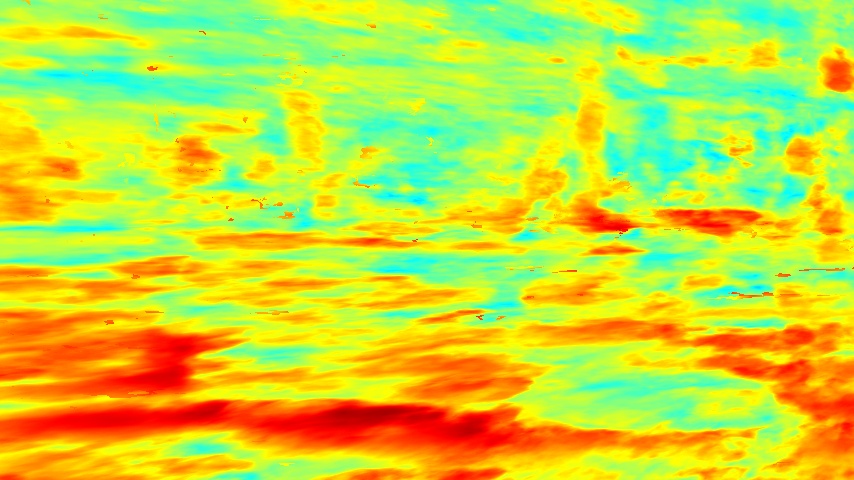} & 
\includegraphics[width=0.17\textwidth]{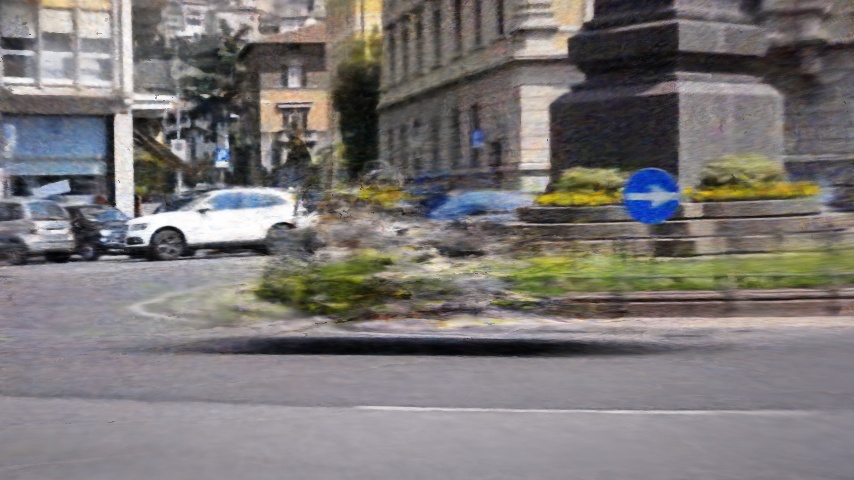} & \includegraphics[width=0.17\textwidth]{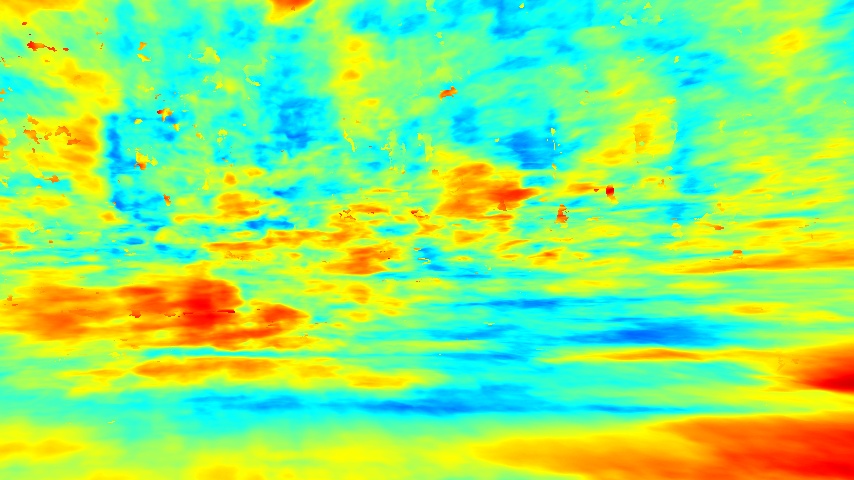} \\
\raisebox{2.2\normalbaselineskip}[0pt][0pt]{\rotatebox[origin=c]{90}{BARF~\cite{lin2021barf}}} &
\includegraphics[width=0.17\textwidth]{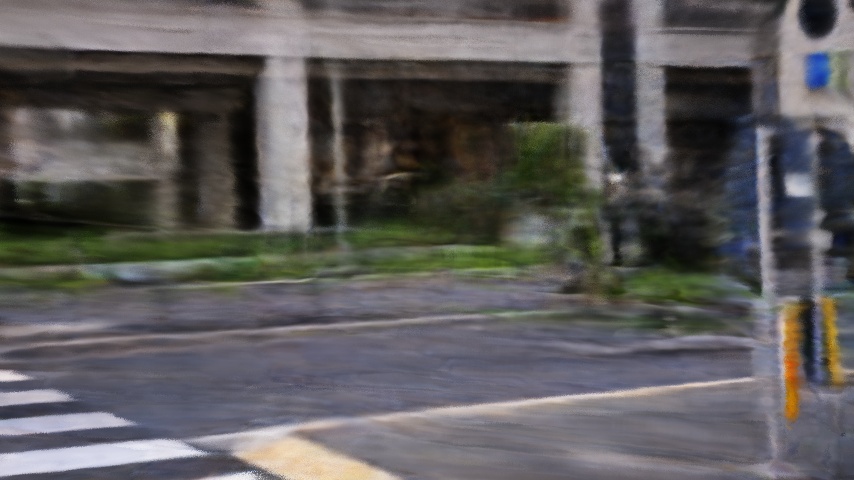} & \includegraphics[width=0.17\textwidth]{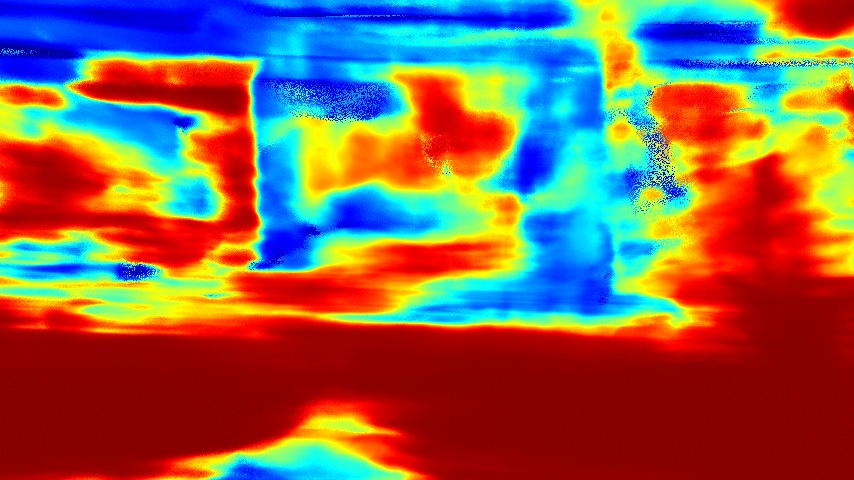} & 
\includegraphics[width=0.17\textwidth]{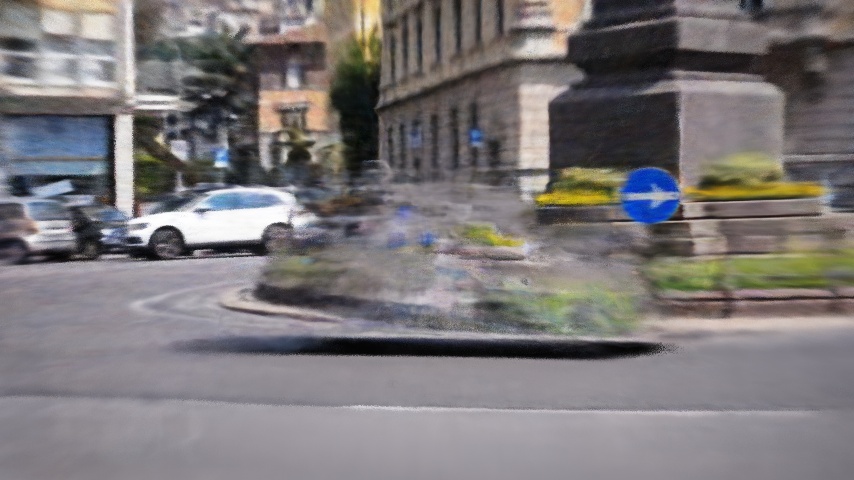} & \includegraphics[width=0.17\textwidth]{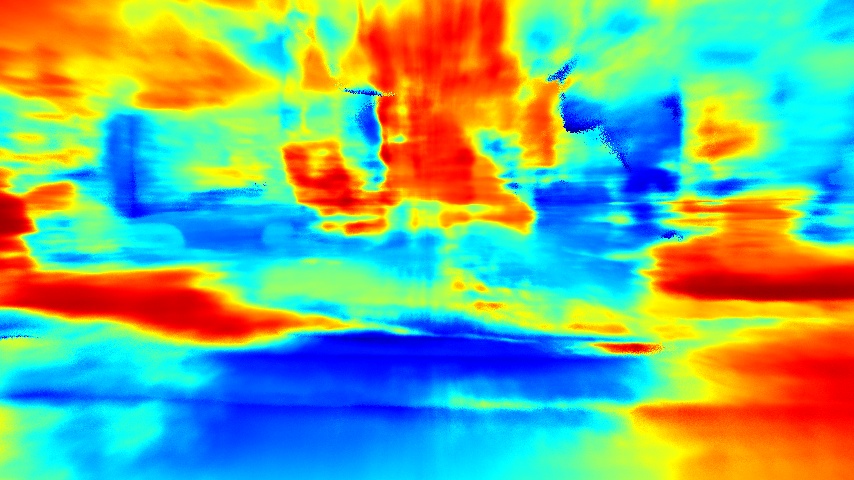} \\
\raisebox{2.2\normalbaselineskip}[0pt][0pt]{\rotatebox[origin=c]{90}{Ours}} &
\includegraphics[width=0.17\textwidth]{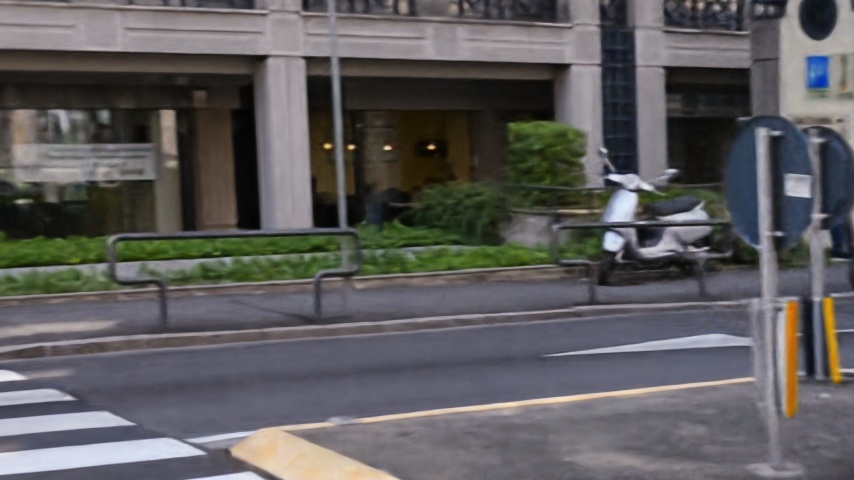} & \includegraphics[width=0.17\textwidth]{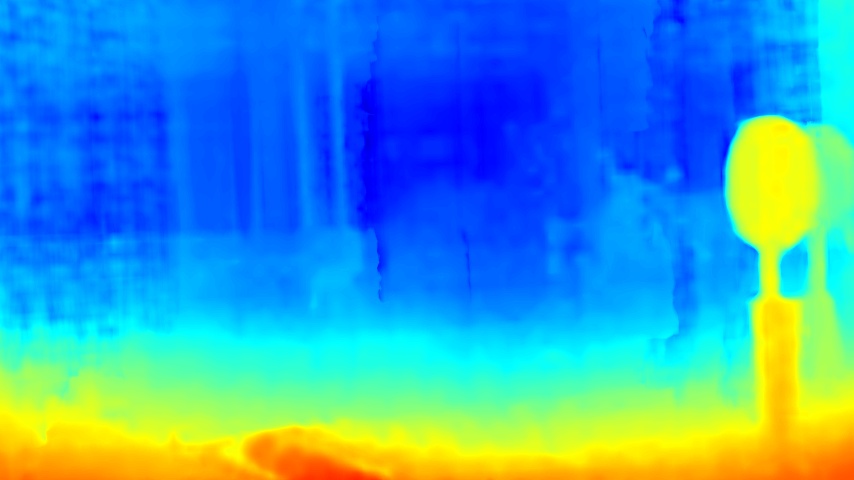} & 
\includegraphics[width=0.17\textwidth]{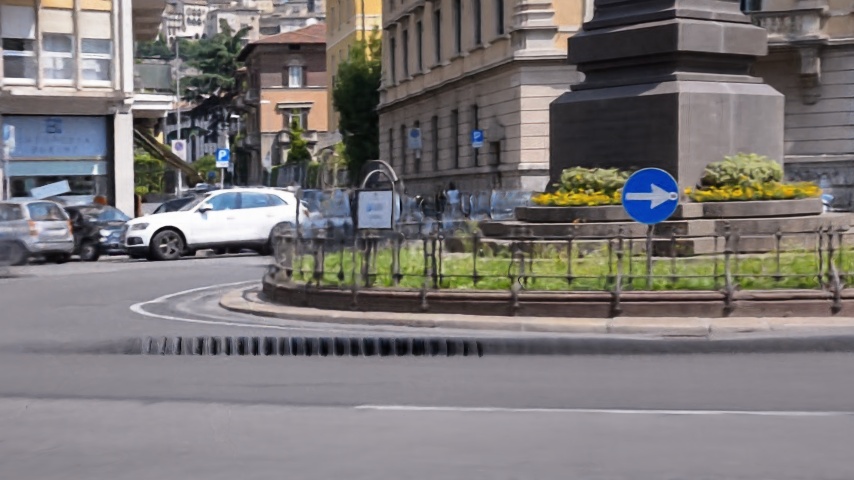} & \includegraphics[width=0.17\textwidth]{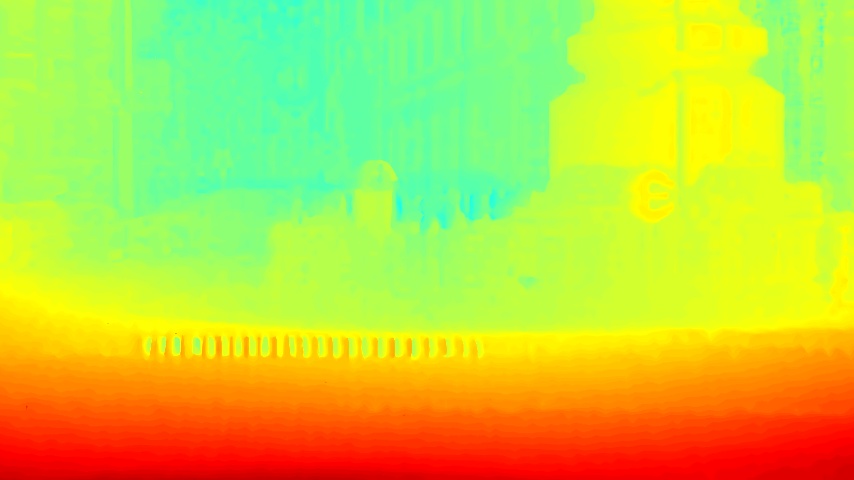} \\
 & Image & Depth & Image & Depth \\\
\end{tabular}%
}
\vspace{-5mm}
\caption{\textbf{Qualitative results of static view synthesis on the DAVIS dataset from \emph{unknown} camera poses and \emph{ground truth} foreground masks.}
}
\label{fig:visual_BARF}
\end{figure}

We then linearly compose the static and dynamic parts into the final results with the predicted nonrigidity $m^d$:
\begin{equation}
\begin{gathered}
\hat{\mathbf{C}}(\mathbf{r}) = \sum_{i=1}^{N} T(i)(m^d(1-\text{exp}(-\sigma^d(i)\delta(i)))\mathbf{c}^d(i)+ \\
(1-m^d)(1-\text{exp}(-\sigma^s(i)\delta(i)))\mathbf{c}^s(i)).
\end{gathered}
\end{equation}

\topic{Total training loss.}
The total training loss is:
\begin{equation}
\mathcal{L} = \left \| \hat{\mathbf{C}}(\mathbf{r}) - \mathbf{C}(\mathbf{r}) \right \|_2^2 + \mathcal{L}^s + \mathcal{L}^d.
\end{equation}

\begin{figure*}[]
\centering
\setlength{\tabcolsep}{1pt}
\renewcommand{\arraystretch}{0.4}
\resizebox{\textwidth}{!}{%
\begin{tabular}{cccccc}
\includegraphics[width=0.17\textwidth]{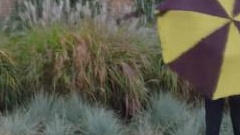} & \includegraphics[width=0.17\textwidth]{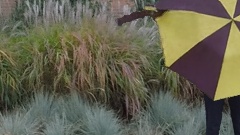} & \includegraphics[width=0.17\textwidth]{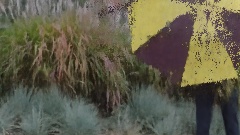} & \includegraphics[width=0.17\textwidth]{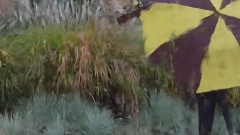} & \includegraphics[width=0.17\textwidth]{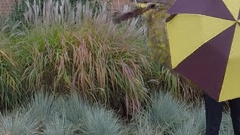} & \includegraphics[width=0.17\textwidth]{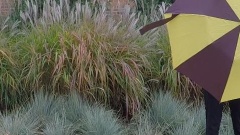} \\
\includegraphics[width=0.17\textwidth]{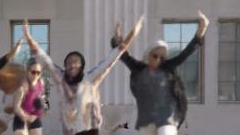} & \includegraphics[width=0.17\textwidth]{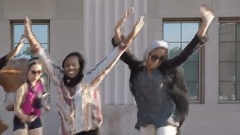} & \includegraphics[width=0.17\textwidth]{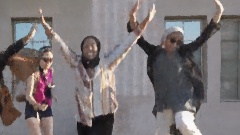} & \includegraphics[width=0.17\textwidth]{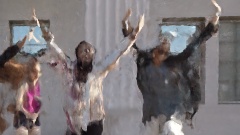} & \includegraphics[width=0.17\textwidth]{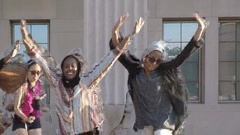} & \includegraphics[width=0.17\textwidth]{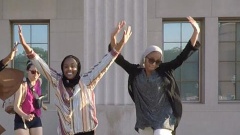} \\
NSFF~\cite{li2021neural} & DynamicNeRF~\cite{gao2021dynamic} & HyperNeRF~\cite{park2021hypernerf} & TiNeuVox~\cite{fang2022fast} & Ours & Ground truth \\
\end{tabular}%
}
\vspace{-2mm}
\caption{\textbf{Novel view synthesis.} Compared to other methods, our results are sharper, closer to the ground truth, and contain fewer artifacts.
}
\label{fig:visual_Nvidia}
\vspace{-2mm}
\end{figure*}

\begin{table*}[t]
\caption{
\textbf{Novel view synthesis results.}
We report the average PSNR and LPIPS results with comparisons to existing methods on Dynamic Scene dataset~\cite{yoon2020novel}. *: Numbers are adopted from DynamicNeRF~\cite{gao2021dynamic}.
}
\vspace{-2mm}
\label{tab:quantitative_nvidia}
\centering
\resizebox{1\linewidth}{!} 
{
\begin{tabular}{l ccccccc|c}
\toprule
PSNR $\uparrow$ / LPIPS $\downarrow$ & Jumping & Skating & Truck & Umbrella & Balloon1 & Balloon2 & Playground & Average \\
\midrule
NeRF*~\cite{mildenhall2020nerf} & 
20.99 / 0.305 &
23.67 / 0.311 &
22.73 / 0.229 &
21.29 / 0.440 &
19.82 / 0.205 &
24.37 / 0.098 &
21.07 / 0.165 &
21.99 / 0.250 \\
%
D-NeRF~\cite{pumarola2021d} & 
22.36 / 0.193 & 
22.48 / 0.323 & 
24.10 / 0.145 & 
21.47 / 0.264 & 
19.06 / 0.259 & 
20.76 / 0.277 & 
20.18 / 0.164 &
21.48 / 0.232 \\
NR-NeRF*~\cite{tretschk2021non} & 
20.09 / 0.287 & 
23.95 / 0.227 & 
19.33 / 0.446 & 
19.63 / 0.421 & 
17.39 / 0.348 & 
22.41 / 0.213 & 
15.06 / 0.317 &
19.69 / 0.323 \\
NSFF*~\cite{li2021neural} & 
24.65 / 0.151 & 
\second{29.29} / 0.129 & 
25.96 / 0.167 & 
22.97 / 0.295 & 
21.96 / 0.215 & 
24.27 / 0.222 & 
21.22 / 0.212 &
24.33 / 0.199 \\
DynamicNeRF*~\cite{gao2021dynamic} & 
\second{24.68} / \second{0.090} & 
\first{32.66} / \first{0.035} & 
\second{28.56} / \second{0.082} & 
\second{23.26} / \second{0.137} &
\second{22.36} / \second{0.104} & 
\first{27.06} / \first{0.049} & 
\second{24.15} / \second{0.080} &
\first{26.10} / \second{0.082} \\
HyperNeRF~\cite{park2021hypernerf} & 
18.34 / 0.302 &
21.97 / 0.183 & 
20.61 / 0.205 & 
18.59 / 0.443 & 
13.96 / 0.530 & 
16.57 / 0.411 & 
13.17 / 0.495 &
17.60 / 0.367 \\
TiNeuVox~\cite{fang2022fast} & 
20.81 / 0.247 &
23.32 / 0.152 & 
23.86 / 0.173 & 
20.00 / 0.355 & 
17.30 / 0.353 & 
19.06 / 0.279 & 
13.84 / 0.437 &
19.74 / 0.285 \\
%
%
%
Ours w/ COLMAP poses & 
\first{25.66} / \first{0.071} & 
28.68 / \second{0.040} & 
\first{29.13} / \first{0.063} & 
\first{24.26} / \first{0.089} & 
\first{22.37} / \first{0.103} & 
\second{26.19} / \second{0.054} & 
\first{24.96} / \first{0.048} & 
\second{25.89} / \first{0.065} \\
\midrule
Ours w/o COLMAP poses & 
24.27 / 0.100 & 
28.71 / 0.046 & 
28.85 / 0.066 & 
23.25 / 0.104 & 
21.81 / 0.122 & 
25.58 / 0.064 & 
25.20 / 0.052 &
25.38 / 0.079 \\
\bottomrule
\end{tabular}
}
\vspace{-2mm}
\end{table*}

\subsection{Implementation Details}
\label{sec:details}
We simultaneously estimate camera poses, focal length, static radiance fields, and dynamic radiance fields. 
For forward-facing scenes, we parameterize the scenes with normalized device coordinates (NDC). 
To handle unbounded scenes in the wild videos, we parameterize the scenes using the contraction parameterization~\cite{barron2022mip}. 
To encourage solid surface scene reconstruction and prevent floaters, we add the distortion loss~\cite{barron2022mip,sun2022direct}. 
We set the finest voxel resolution to 262,144,000 and 27,000,000 for NDC and contraction, respectively. 
We also decompose the voxel grid using the VM-decomposition in TensoRF~\cite{chen2022tensorf} for model compactness. 
The entire training process takes around 28 hours with one NVIDIA V100 GPU. 
We provide the detailed architecture in the supplementary material.

\section{Experimental Results}
\label{sec:experimental_results}
Due to the space limit, we leave the experimental setup, including datasets, compared methods, and the evaluation metrics to the supplementary materials.

\begin{table*}[t]
\caption{
\textbf{Novel view synthesis results.}
We compare the mPSNR and mSSIM scores with existing methods on the iPhone dataset~\cite{gao2022monocular}.
}
\vspace{-2mm}
\label{tab:quantitative_iphone}
\centering
\resizebox{1\linewidth}{!} 
{
\begin{tabular}{l ccccccc|c}
\toprule
%
mPSNR $\uparrow$ / mSSIM $\uparrow$ & Apple & Block & Paper-windmill & Space-out & Spin & Teddy & Wheel & Average \\
\midrule
NSFF~\cite{li2021neural} & 17.54 / 0.750 & 16.61 / 0.639 & 17.34 / 0.378 & 17.79 / 0.622 & 18.38 / 0.585 & 13.65 / 0.557 & 13.82 / 0.458 & 15.46 / 0.569 \\
Nerfies~\cite{park2021nerfies} & 17.64 / 0.743 & 17.54 / 0.670 & 17.38 / 0.382 & 17.93 / 0.605 & 19.20 / 0.561 & 13.97 / 0.568 & 13.99 / 0.455 & 16.45 / 0.569 \\
HyperNeRF~\cite{park2021hypernerf} & 16.47 / 0.754 & 14.71 / 0.606 & 14.94 / 0.272 & 17.65 / 0.636 & 17.26 / 0.540 & 12.59 / 0.537 & 14.59 / 0.511 & 16.81 / 0.550 \\
T-NeRF~\cite{gao2022monocular} & 17.43 / 0.728 & 17.52 / 0.669 & 17.55 / 0.367 & 17.71 / 0.591 & 19.16 / 0.567 & 13.71 / 0.570 & 15.65 / 0.548 & 16.96 / 0.577 \\
Ours & 18.73 / 0.722 & 18.73 / 0.634 & 16.71 / 0.321 & 18.56 / 0.594 & 17.41 / 0.484 & 14.33 / 0.536 & 15.20 / 0.449 & 17.09 / 0.534 \\
\bottomrule
\end{tabular}
}
\end{table*}
\begin{figure*}[]
\centering
\setlength{\tabcolsep}{1pt}
\renewcommand{\arraystretch}{0.4}
\resizebox{\textwidth}{!}{%
\begin{tabular}{ccccc}
\includegraphics[width=0.2\textwidth]{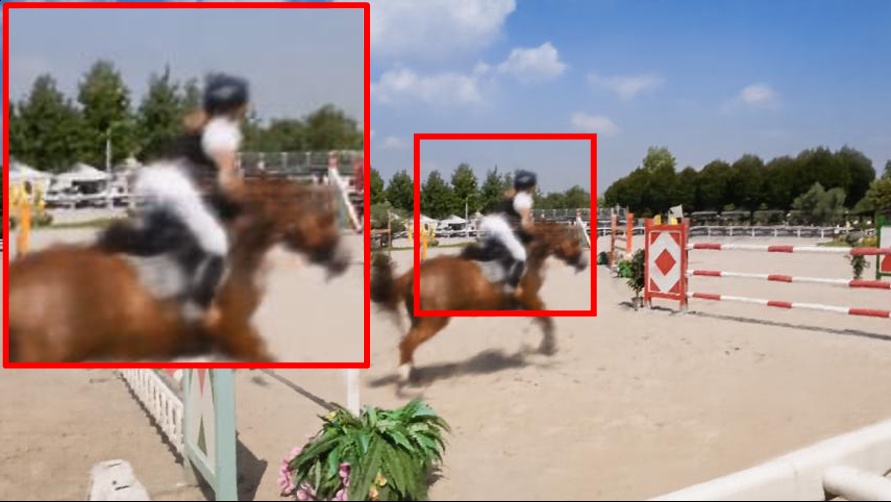} & \includegraphics[width=0.2\textwidth]{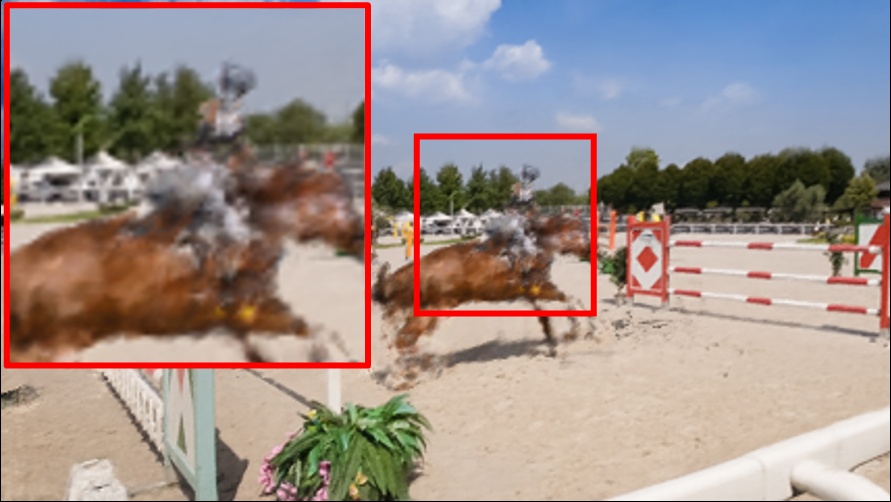} & 
\includegraphics[trim=0 1px 0 1px, clip=true, width=0.2\textwidth]{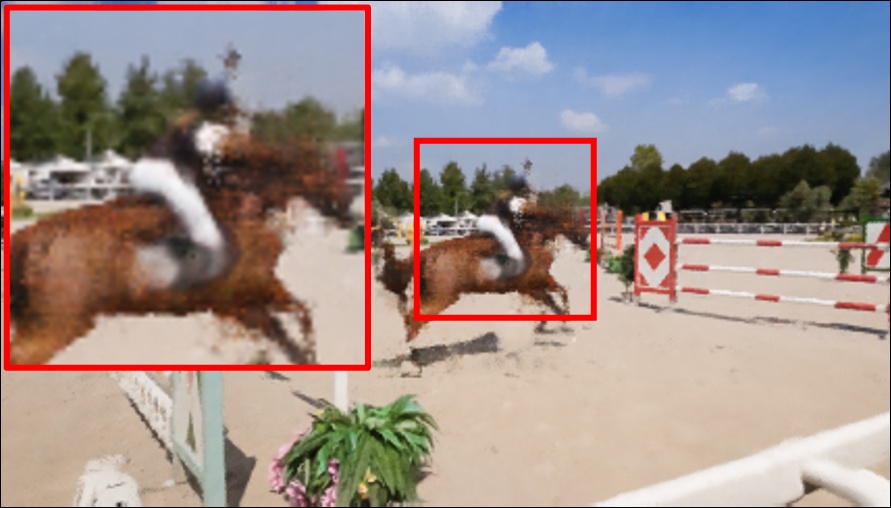} & \includegraphics[width=0.2\textwidth]{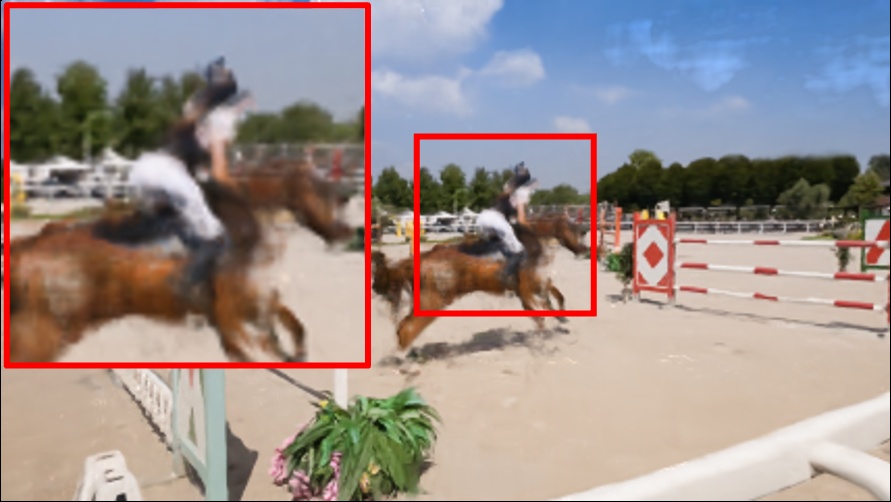} &
\includegraphics[width=0.2\textwidth]{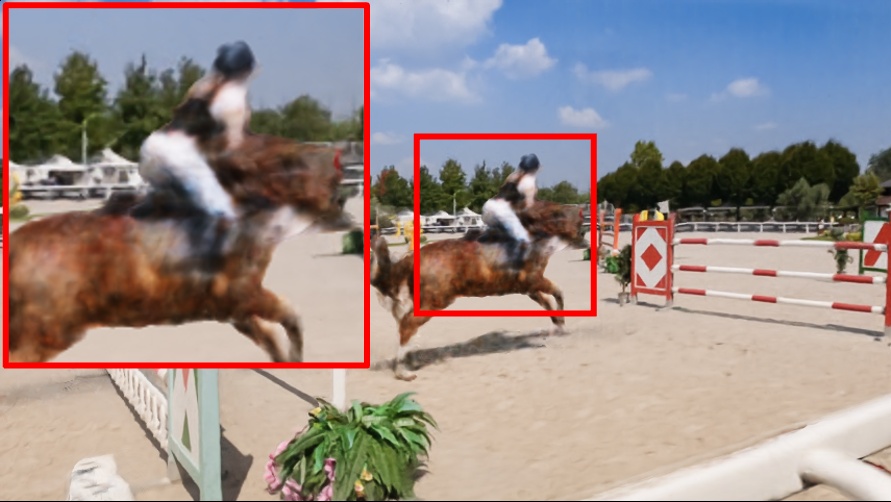} \\
\includegraphics[width=0.2\textwidth]{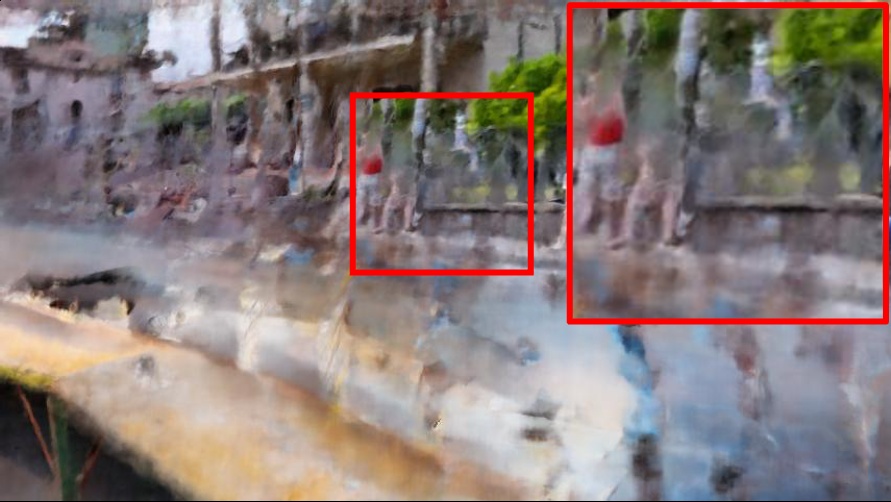} & \includegraphics[width=0.2\textwidth]{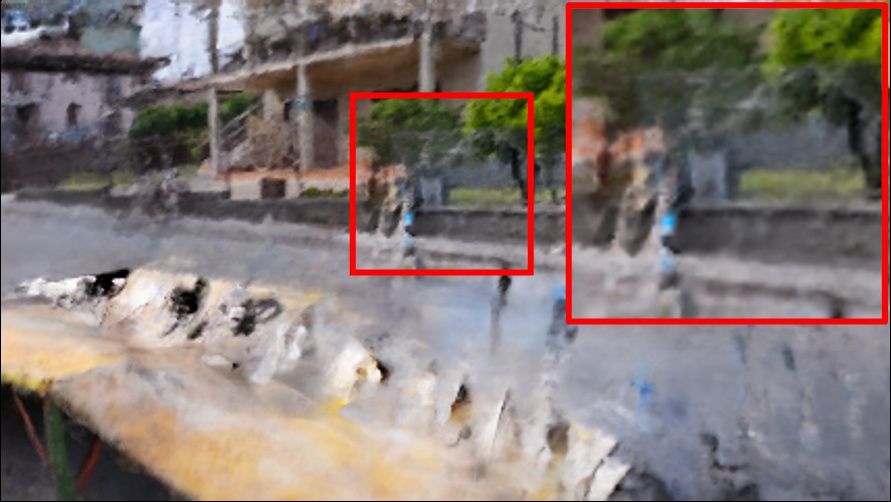} & 
\includegraphics[trim=0 1px 0 1px, clip=true, width=0.2\textwidth]{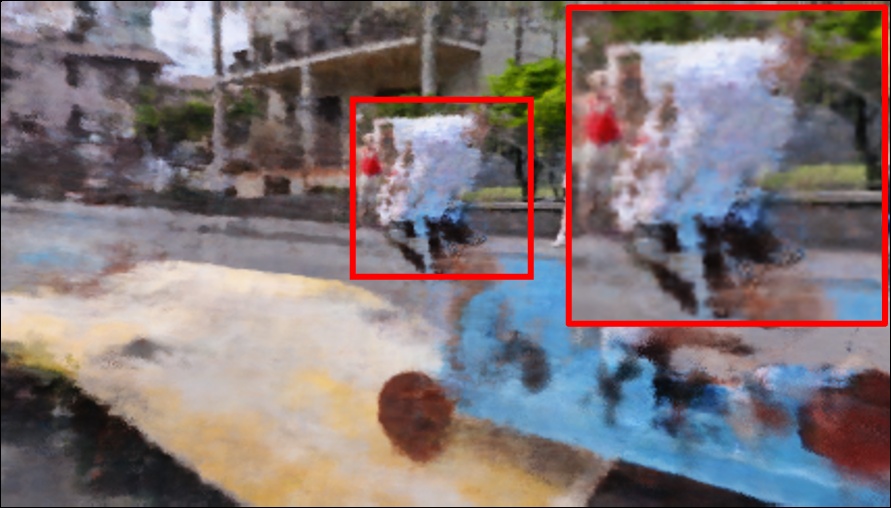} & \includegraphics[width=0.2\textwidth]{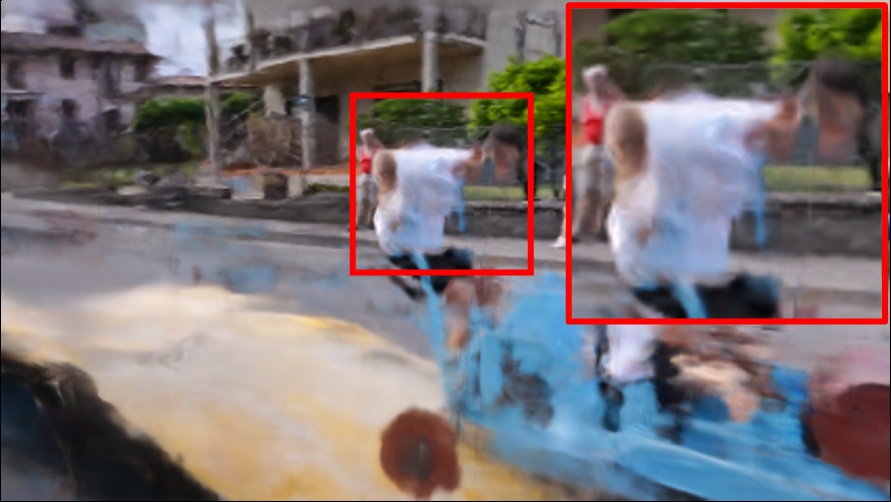} &
\includegraphics[width=0.2\textwidth]{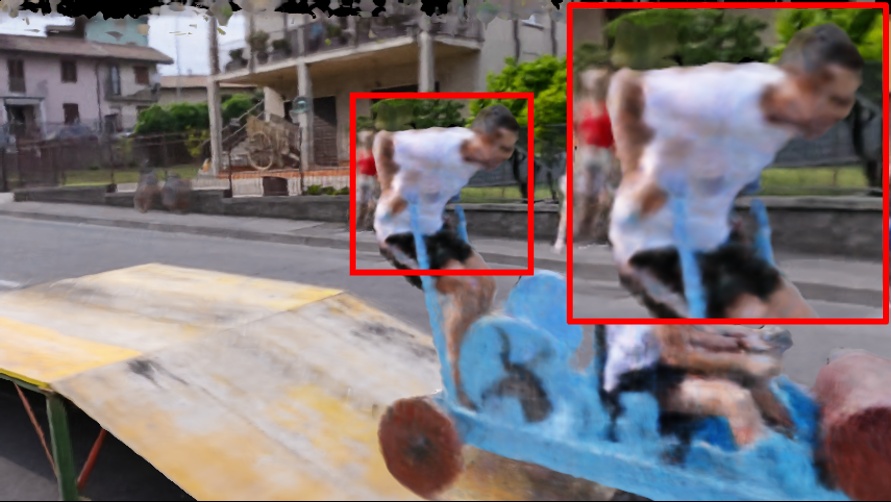} \\
\includegraphics[width=0.2\textwidth]{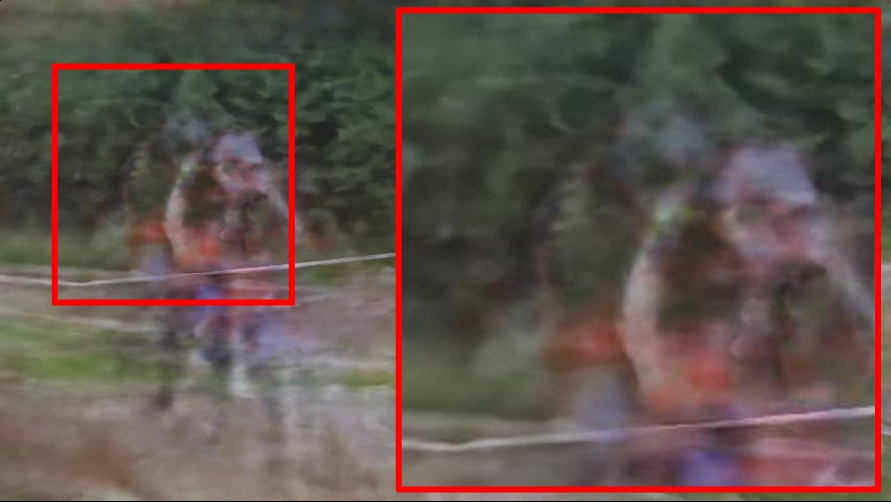} & \includegraphics[width=0.2\textwidth]{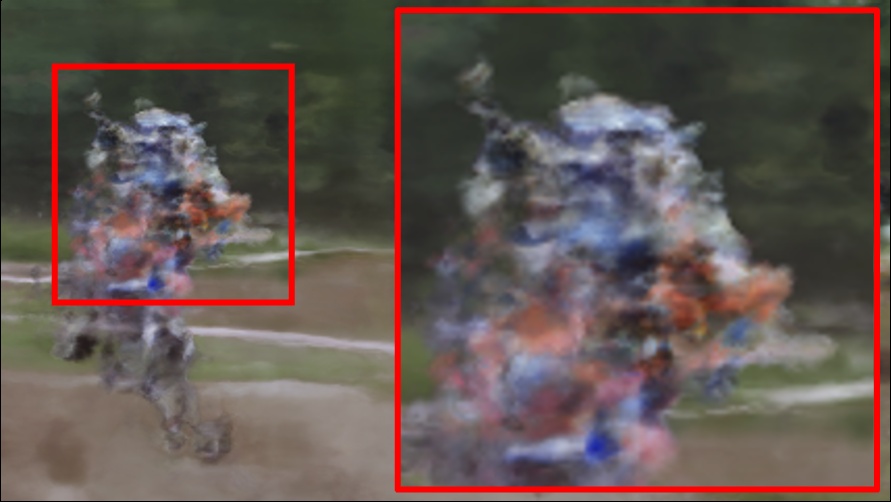} & 
\includegraphics[trim=0 1px 0 1px, clip=true, width=0.2\textwidth]{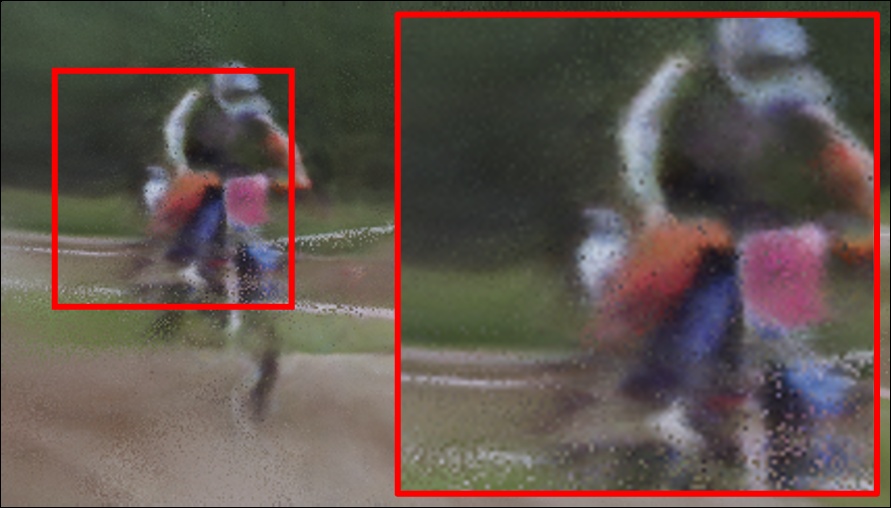} & \includegraphics[width=0.2\textwidth]{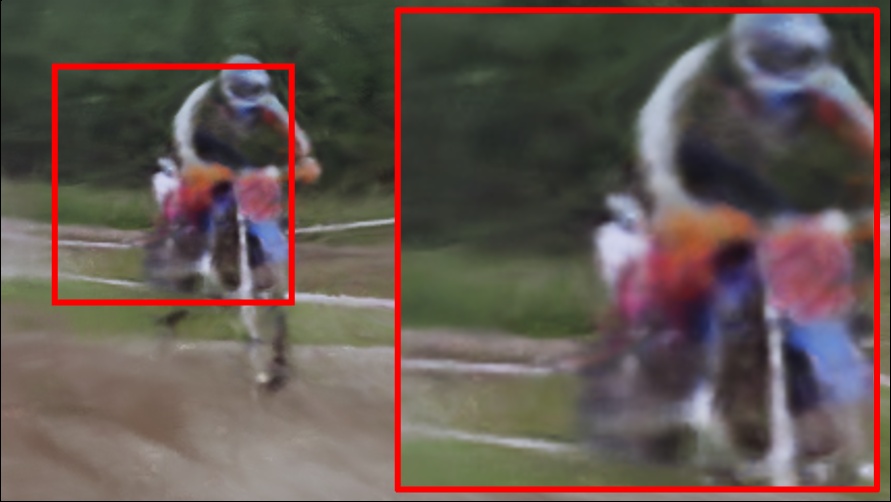} &
\includegraphics[trim=5px 0 5px 0, clip=true, width=0.2\textwidth]{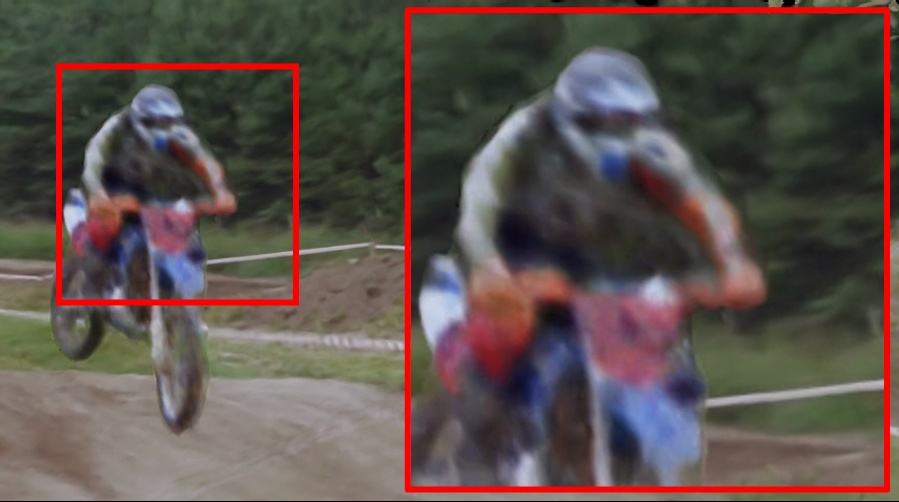} \\
NSFF~\cite{li2021neural} & DynamicNeRF~\cite{gao2021dynamic} & HyperNeRF~\cite{park2021hypernerf} & TiNeuVox~\cite{fang2022fast} & Ours \\
\end{tabular}%
}
\vspace{-2mm}
\caption{\textbf{Novel space-time synthesis results on the DAVIS dataset with \emph{our estimated camera poses}.} COLMAP fails to produce reliable camera poses for most of the sequences in the DAVIS dataset. With the estimated camera poses by our method, we can run other methods and perform space-time synthesis on the scenes that are not feasible with COLMAP. Our method produces images with much better quality.}
\label{fig:visual_DAVIS}
\vspace{-2mm}
\end{figure*}

\subsection{Evaluation on Camera Poses Estimation}
We conduct the camera pose estimation evaluation on the MPI Sintel dataset~\cite{butler2012naturalistic} and show the quantitative results in~\tabref{quantitative_sintel}. 
Our method performs significantly better than existing NeRF-based pose estimation methods. 
Note that our method also performs favorably against existing learning-based visual odometry methods. 
We show some visual comparisons of the predicted camera trajectories in~\figref{visual_Sintel}, and the sorted error plots that show both the \emph{accuracy} and \emph{completeness/robustness} in~\figref{curve_Sintel}. 
Our approach predicts accurate camera poses over other NeRF-based pose estimation methods. 
Our method is a global optimization over the entire sequence instead of local registration like SLAM-based methods. Therefore, our RPE trans and rot scores are slightly worse than ParticleSfM~\cite{zhao2022particlesfm} as consecutive frames' rotation is less accurate.

To further reduce the effect of the dynamic parts, we use the ground truth motion masks provided by the DAVIS dataset to mask out the loss calculations in the dynamic regions for all the NeRF-based compared methods. 
We show the reconstructed images and depth maps in~\figref{visual_BARF}. 
Our approach can successfully reconstruct the detailed content and the faithful geometry thanks to the auxiliary losses.
On the contrary, other methods often fail to reconstruct consistent scene geometry and thus produce poor synthesis results.

\subsection{Evaluation on Dynamic View Synthesis}
\topic{Quantitative evaluation.}
We follow the evaluation protocol in DynamicNeRF~\cite{gao2021dynamic} to synthesize the view from the first camera and change time on the NVIDIA dynamic view synthesis dataset. 
We report the PSNR and LPIPS in~\tabref{quantitative_nvidia}.
Our method performs favorably against state-of-the-art methods. 
Furthermore, even without COLMAP poses, our method can still achieve results comparable to the ones using COLMAP poses.

We also follow the evaluation protocol in DyCheck~\cite{gao2022monocular} and evaluate quantitatively on the iPhone dataset~\cite{gao2022monocular}. 
We report the masked PSNR and SSIM in~\tabref{quantitative_iphone} and show that our method performs on par with existing methods.

\topic{Qualitative evaluation.}
We show some visual comparisons on the NVIDIA dynamic view synthesis dataset in~\figref{visual_Nvidia} and DAVIS dataset in~\figref{visual_DAVIS}. 
COLMAP fails to estimate the camera poses for 44 out of 50 sequences in the DAVIS dataset. 
Therefore, we first run our method and give our camera poses to other methods as input. 
With the joint learning of the camera poses and radiance fields, our method produces frames with fewer visual artifacts. Other methods can also benefit from our estimated poses to synthesize novel views. 
With our poses, they can reconstruct consistent static scenes but often generate artifacts for the dynamic parts. 
In contrast, our method utilizes the auxiliary priors and thus produces results of much better visual quality.

\begin{table}[t]
    \caption{
    \textbf{Ablation studies.}
    We report PSNR, SSIM and LPIPS on the Playground sequence.
    }
    \vspace{-2mm}
    \label{tab:ablation}
    \centering
    \small{(a) Pose estimation design choices}
    \resizebox{1.0\linewidth}{!} 
    {
    \begin{tabular}{l | ccc}
    \toprule
    & PSNR $\uparrow$ & SSIM $\uparrow$ & LPIPS $\downarrow$ \\
    \midrule
    Ours w/o coarse-to-fine                   & 12.45 & 0.4829 & 0.327 \\
    Ours w/o late viewing direction fusion    & 18.34 & 0.5521 & 0.263 \\
    Ours w/o stopping the dynamic gradients   & 21.47 & 0.7392 & 0.211 \\
    Ours                                      & \textbf{25.20} & \textbf{0.9052} & \textbf{0.052} \\
    \bottomrule
    \end{tabular}
    }
    \small{(b) Dynamic reconstruction achitectural designs}
    \resizebox{1.0\linewidth}{!} 
    {
    \begin{tabular}{ccc|ccc}
    \toprule
    Dyn. model & Deform. MLP & Time-depend. MLPs & PSNR $\uparrow$ & SSIM $\uparrow$ & LPIPS $\downarrow$ \\
    \midrule
    & & & 21.34 & 0.8192 & 0.161 \\
    \checkmark & \checkmark & & 22.37 & 0.8317 & 0.115 \\
    \checkmark & & \checkmark & 23.14 & 0.8683 & 0.083 \\
    \checkmark & \checkmark & \checkmark & \textbf{25.20} & \textbf{0.9052} & \textbf{0.052} \\
    %
    \bottomrule
    \end{tabular}
    }
\vspace{-2mm}
\end{table}

\subsection{Ablation Study}


We analyze the design choices in~\tabref{ablation}.
For the camera poses estimation, the coarse-to-fine voxel upsampling strategy is the most critical component. 
Late viewing direction fusion and stopping the gradients from the dynamic radiance field also help the optimization find better poses and lead to higher-quality rendering results. 
Please refer to~\figref{pose_ablation} for visual comparisons.
For the dynamic radiance field reconstruction, both the deformation MLP and the time-dependent MLPs improve the final rendering quality.

\begin{figure}[]
\centering
\setlength{\tabcolsep}{1pt}
\renewcommand{\arraystretch}{1}
\resizebox{\columnwidth}{!}{%
\begin{tabular}{cccc}
\includegraphics[height=0.2\columnwidth]{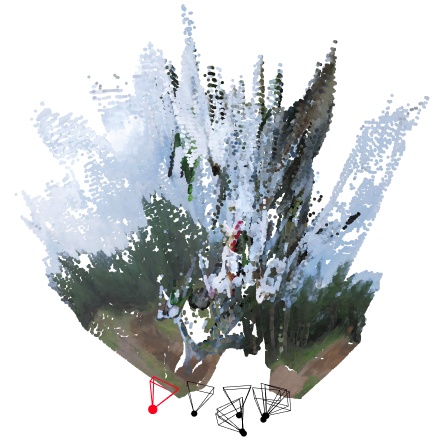} & \includegraphics[height=0.2\columnwidth]{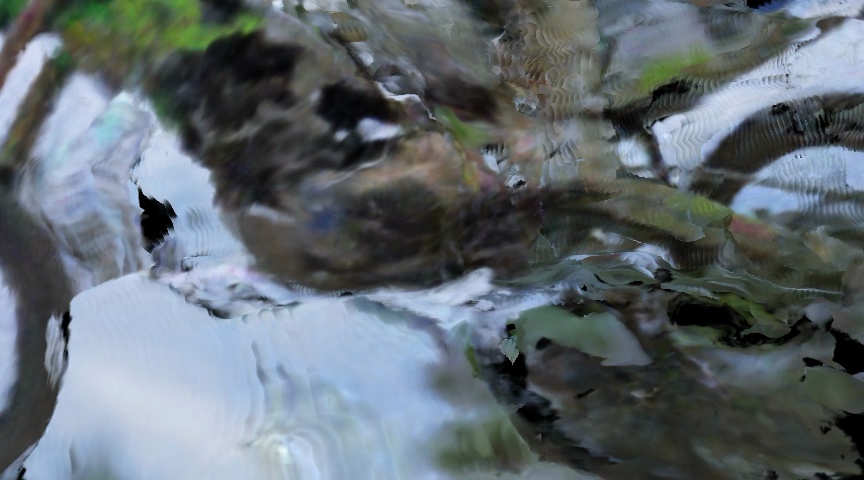} & \includegraphics[height=0.2\columnwidth]{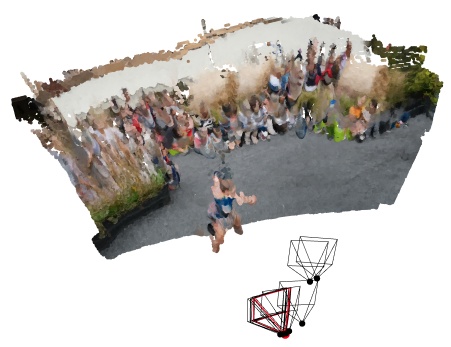} & \includegraphics[height=0.2\columnwidth]{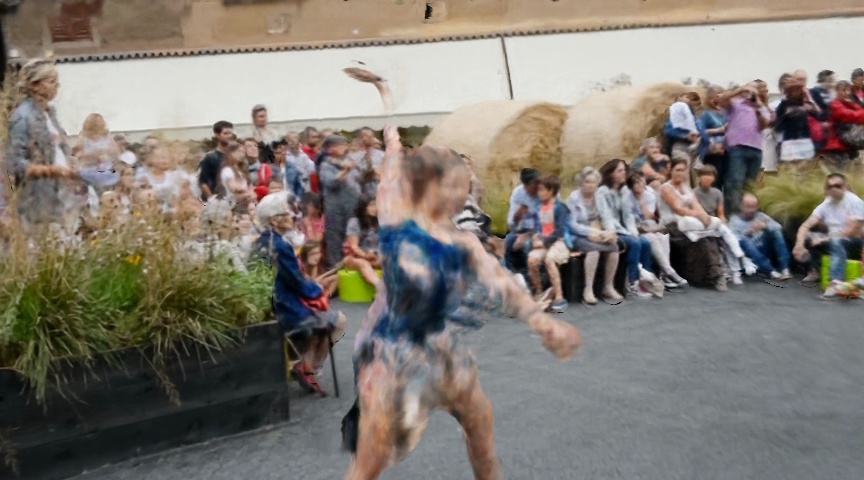} \\
\multicolumn{2}{c}{(a) Fast moving camera} & \multicolumn{2}{c}{(b) Changing focal length} \\
\end{tabular}%
}
\vspace{-2mm}
\caption{\textbf{Failure cases.} (a) In the cases that the camera is moving fast, the flow estimation fails and leads to wrong estimated poses and geometry. (b) Our method assumes a shared intrinsic over the entire video and thus cannot handle changing focal length well.}
\label{fig:failure}
\vspace{-2mm}
\end{figure}


\subsection{Failure Cases}
Even with these efforts, robust dynamic view synthesis from a monocular video without known camera poses is still challenging. We show some failure cases in~\figref{failure}.


\section{Conclusions}
\label{sec:conclusions}

We present \emph{robust dynamic radiance fields} for space-time synthesis of casually captured monocular videos without requiring camera poses as input. 
With the proposed model designs, we demonstrate that our approach can reconstruct accurate dynamic radiance fields from a wide range of challenging videos. 
We validate the efficacy of the proposed method via extensive quantitative and qualitative comparisons with the state-of-the-art. 



{\small
\bibliographystyle{ieee_fullname}
\bibliography{egbib}

\begin{thebibliography}{10}\itemsep=-1pt

\bibitem{ballan2010unstructured}
Luca Ballan, Gabriel~J Brostow, Jens Puwein, and Marc Pollefeys.
\newblock Unstructured video-based rendering: Interactive exploration of
  casually captured videos.
\newblock {\em ACM TOG}, pages 1--11, 2010.

\bibitem{ballester2021dot}
Irene Ballester, Alejandro Font{\'a}n, Javier Civera, Klaus~H Strobl, and
  Rudolph Triebel.
\newblock Dot: Dynamic object tracking for visual slam.
\newblock In {\em 2021 IEEE International Conference on Robotics and Automation
  (ICRA)}, 2021.

\bibitem{bansal20204d}
Aayush Bansal, Minh Vo, Yaser Sheikh, Deva Ramanan, and Srinivasa Narasimhan.
\newblock 4d visualization of dynamic events from unconstrained multi-view
  videos.
\newblock In {\em CVPR}, 2020.

\bibitem{barron2021mip}
Jonathan~T Barron, Ben Mildenhall, Matthew Tancik, Peter Hedman, Ricardo
  Martin-Brualla, and Pratul~P Srinivasan.
\newblock Mip-nerf: A multiscale representation for anti-aliasing neural
  radiance fields.
\newblock In {\em ICCV}, 2021.

\bibitem{barron2022mip}
Jonathan~T Barron, Ben Mildenhall, Dor Verbin, Pratul~P Srinivasan, and Peter
  Hedman.
\newblock Mip-nerf 360: Unbounded anti-aliased neural radiance fields.
\newblock In {\em CVPR}, 2022.

\bibitem{bescos2018dynaslam}
Berta Bescos, Jos{\'e}~M F{\'a}cil, Javier Civera, and Jos{\'e} Neira.
\newblock Dynaslam: Tracking, mapping, and inpainting in dynamic scenes.
\newblock {\em IEEE Robotics and Automation Letters}, 2018.

\bibitem{broxton2020immersive}
Michael Broxton, John Flynn, Ryan Overbeck, Daniel Erickson, Peter Hedman,
  Matthew Duvall, Jason Dourgarian, Jay Busch, Matt Whalen, and Paul Debevec.
\newblock Immersive light field video with a layered mesh representation.
\newblock {\em ACM TOG}, 39:86--1, 2020.

\bibitem{buehler2001unstructured}
Chris Buehler, Michael Bosse, Leonard McMillan, Steven Gortler, and Michael
  Cohen.
\newblock Unstructured lumigraph rendering.
\newblock In {\em Proceedings of the 28th annual conference on Computer
  graphics and interactive techniques}, pages 425--432, 2001.

\bibitem{butler2012naturalistic}
Daniel~J Butler, Jonas Wulff, Garrett~B Stanley, and Michael~J Black.
\newblock A naturalistic open source movie for optical flow evaluation.
\newblock In {\em ECCV}, 2012.

\bibitem{carranza2003free}
Joel Carranza, Christian Theobalt, Marcus~A Magnor, and Hans-Peter Seidel.
\newblock Free-viewpoint video of human actors.
\newblock {\em ACM TOG}, 22:569--577, 2003.

\bibitem{chaurasia2013depth}
Gaurav Chaurasia, Sylvain Duchene, Olga Sorkine-Hornung, and George Drettakis.
\newblock Depth synthesis and local warps for plausible image-based navigation.
\newblock {\em ACM TOG}, 32:1--12, 2013.

\bibitem{chen2022tensorf}
Anpei Chen, Zexiang Xu, Andreas Geiger, Jingyi Yu, and Hao Su.
\newblock Tensorf: Tensorial radiance fields.
\newblock In {\em ECCV}, 2022.

\bibitem{chen1993view}
Shenchang~Eric Chen and Lance Williams.
\newblock View interpolation for image synthesis.
\newblock In {\em Proceedings of the 20th annual conference on Computer
  graphics and interactive techniques}, pages 279--288, 1993.

\bibitem{choi2019extreme}
Inchang Choi, Orazio Gallo, Alejandro Troccoli, Min~H Kim, and Jan Kautz.
\newblock Extreme view synthesis.
\newblock In {\em ICCV}, 2019.

\bibitem{collet2015high}
Alvaro Collet, Ming Chuang, Pat Sweeney, Don Gillett, Dennis Evseev, David
  Calabrese, Hugues Hoppe, Adam Kirk, and Steve Sullivan.
\newblock High-quality streamable free-viewpoint video.
\newblock {\em ACM TOG}, 34:1--13, 2015.

\bibitem{dou2016fusion4d}
Mingsong Dou, Sameh Khamis, Yury Degtyarev, Philip Davidson, Sean~Ryan Fanello,
  Adarsh Kowdle, Sergio~Orts Escolano, Christoph Rhemann, David Kim, Jonathan
  Taylor, et~al.
\newblock Fusion4d: Real-time performance capture of challenging scenes.
\newblock {\em ACM TOG}, 35:1--13, 2016.

\bibitem{drebin1988volume}
Robert~A Drebin, Loren Carpenter, and Pat Hanrahan.
\newblock Volume rendering.
\newblock {\em ACM TOG}, 22:65--74, 1988.

\bibitem{engel2017direct}
Jakob Engel, Vladlen Koltun, and Daniel Cremers.
\newblock Direct sparse odometry.
\newblock {\em IEEE TPAMI}, 40:611--625, 2017.

\bibitem{engel2014lsd}
Jakob Engel, Thomas Sch{\"o}ps, and Daniel Cremers.
\newblock Lsd-slam: Large-scale direct monocular slam.
\newblock In {\em ECCV}, 2014.

\bibitem{fang2022fast}
Jiemin Fang, Taoran Yi, Xinggang Wang, Lingxi Xie, Xiaopeng Zhang, Wenyu Liu,
  Matthias Nie{\ss}ner, and Qi Tian.
\newblock Fast dynamic radiance fields with time-aware neural voxels.
\newblock {\em ACM TOG}, 2022.

\bibitem{flynn2019deepview}
John Flynn, Michael Broxton, Paul Debevec, Matthew DuVall, Graham Fyffe, Ryan
  Overbeck, Noah Snavely, and Richard Tucker.
\newblock Deepview: View synthesis with learned gradient descent.
\newblock In {\em CVPR}, 2019.

\bibitem{flynn2016deepstereo}
John Flynn, Ivan Neulander, James Philbin, and Noah Snavely.
\newblock Deepstereo: Learning to predict new views from the world's imagery.
\newblock In {\em CVPR}, 2016.

\bibitem{fridovich2022plenoxels}
Sara Fridovich-Keil, Alex Yu, Matthew Tancik, Qinhong Chen, Benjamin Recht, and
  Angjoo Kanazawa.
\newblock Plenoxels: Radiance fields without neural networks.
\newblock In {\em CVPR}, 2022.

\bibitem{gao2021dynamic}
Chen Gao, Ayush Saraf, Johannes Kopf, and Jia-Bin Huang.
\newblock Dynamic view synthesis from dynamic monocular video.
\newblock In {\em ICCV}, 2021.

\bibitem{gao2022monocular}
Hang Gao, Ruilong Li, Shubham Tulsiani, Bryan Russell, and Angjoo Kanazawa.
\newblock Monocular dynamic view synthesis: A reality check.
\newblock In {\em NeurIPS}, 2022.

\bibitem{godard2019digging}
Cl{\'e}ment Godard, Oisin Mac~Aodha, Michael Firman, and Gabriel~J Brostow.
\newblock Digging into self-supervised monocular depth estimation.
\newblock In {\em ICCV}, 2019.

\bibitem{gortler1996lumigraph}
Steven~J Gortler, Radek Grzeszczuk, Richard Szeliski, and Michael~F Cohen.
\newblock The lumigraph.
\newblock In {\em Proceedings of the 23rd annual conference on Computer
  graphics and interactive techniques}, pages 43--54, 1996.

\bibitem{habermann2019livecap}
Marc Habermann, Weipeng Xu, Michael Zollhoefer, Gerard Pons-Moll, and Christian
  Theobalt.
\newblock Livecap: Real-time human performance capture from monocular video.
\newblock {\em ACM TOG}, 38:1--17, 2019.

\bibitem{he2017mask}
Kaiming He, Georgia Gkioxari, Piotr Doll{\'a}r, and Ross Girshick.
\newblock Mask r-cnn.
\newblock In {\em ICCV}, 2017.

\bibitem{hedman2018deep}
Peter Hedman, Julien Philip, True Price, Jan-Michael Frahm, George Drettakis,
  and Gabriel Brostow.
\newblock Deep blending for free-viewpoint image-based rendering.
\newblock {\em ACM TOG}, 37:1--15, 2018.

\bibitem{jeong2021self}
Yoonwoo Jeong, Seokjun Ahn, Christopher Choy, Anima Anandkumar, Minsu Cho, and
  Jaesik Park.
\newblock Self-calibrating neural radiance fields.
\newblock In {\em ICCV}, 2021.

\bibitem{kajiya1984ray}
James~T Kajiya and Brian~P Von~Herzen.
\newblock Ray tracing volume densities.
\newblock {\em ACM TOG}, 18:165--174, 1984.

\bibitem{kopf2014first}
Johannes Kopf, Michael~F Cohen, and Richard Szeliski.
\newblock First-person hyper-lapse videos.
\newblock {\em ACM TOG}, 33:1--10, 2014.

\bibitem{kopf2020one}
Johannes Kopf, Kevin Matzen, Suhib Alsisan, Ocean Quigley, Francis Ge, Yangming
  Chong, Josh Patterson, Jan-Michael Frahm, Shu Wu, Matthew Yu, et~al.
\newblock One shot 3d photography.
\newblock {\em ACM TOG}, 39:76--1, 2020.

\bibitem{kopf2021robust}
Johannes Kopf, Xuejian Rong, and Jia-Bin Huang.
\newblock Robust consistent video depth estimation.
\newblock In {\em CVPR}, 2021.

\bibitem{kumar2017monocular}
Suryansh Kumar, Yuchao Dai, and Hongdong Li.
\newblock Monocular dense 3d reconstruction of a complex dynamic scene from two
  perspective frames.
\newblock In {\em ICCV}, 2017.

\bibitem{levoy1996light}
Marc Levoy and Pat Hanrahan.
\newblock Light field rendering.
\newblock In {\em Proceedings of the 23rd annual conference on Computer
  graphics and interactive techniques}, pages 31--42, 1996.

\bibitem{li2019learning}
Zhengqi Li, Tali Dekel, Forrester Cole, Richard Tucker, Noah Snavely, Ce Liu,
  and William~T Freeman.
\newblock Learning the depths of moving people by watching frozen people.
\newblock In {\em CVPR}, 2019.

\bibitem{li2021neural}
Zhengqi Li, Simon Niklaus, Noah Snavely, and Oliver Wang.
\newblock Neural scene flow fields for space-time view synthesis of dynamic
  scenes.
\newblock In {\em CVPR}, 2021.

\bibitem{lin2021barf}
Chen-Hsuan Lin, Wei-Chiu Ma, Antonio Torralba, and Simon Lucey.
\newblock Barf: Bundle-adjusting neural radiance fields.
\newblock In {\em ICCV}, 2021.

\bibitem{lin2021deep}
Kai-En Lin, Lei Xiao, Feng Liu, Guowei Yang, and Ravi Ramamoorthi.
\newblock Deep 3d mask volume for view synthesis of dynamic scenes.
\newblock In {\em ICCV}, 2021.

\bibitem{liu2021hybrid}
Yu-Lun Liu, Wei-Sheng Lai, Ming-Hsuan Yang, Yung-Yu Chuang, and Jia-Bin Huang.
\newblock Hybrid neural fusion for full-frame video stabilization.
\newblock In {\em ICCV}, 2021.

\bibitem{lombardi2019neural}
Stephen Lombardi, Tomas Simon, Jason Saragih, Gabriel Schwartz, Andreas
  Lehrmann, and Yaser Sheikh.
\newblock Neural volumes: Learning dynamic renderable volumes from images.
\newblock {\em ACM TOG}, 38:65:1--65:14, 2019.

\bibitem{mildenhall2020nerf}
Ben Mildenhall, Pratul~P. Srinivasan, Matthew Tancik, Jonathan~T. Barron, Ravi
  Ramamoorthi, and Ren Ng.
\newblock Ne{RF}: Representing scenes as neural radiance fields for view
  synthesis.
\newblock In {\em ECCV}, 2020.

\bibitem{muller2022instant}
Thomas M{\"u}ller, Alex Evans, Christoph Schied, and Alexander Keller.
\newblock Instant neural graphics primitives with a multiresolution hash
  encoding.
\newblock {\em ACM TOG}, 41:102:1--102:15, 2022.

\bibitem{mur2015orb}
Raul Mur-Artal, Jose Maria~Martinez Montiel, and Juan~D Tardos.
\newblock Orb-slam: a versatile and accurate monocular slam system.
\newblock {\em IEEE transactions on robotics}, 31:1147--1163, 2015.

\bibitem{mur2017orb}
Raul Mur-Artal and Juan~D Tard{\'o}s.
\newblock Orb-slam2: An open-source slam system for monocular, stereo, and
  rgb-d cameras.
\newblock {\em IEEE transactions on robotics}, 33:1255--1262, 2017.

\bibitem{newcombe2011dtam}
Richard~A Newcombe, Steven~J Lovegrove, and Andrew~J Davison.
\newblock {DTAM}: Dense tracking and mapping in real-time.
\newblock In {\em ICCV}, 2011.

\bibitem{niklaus20193d}
Simon Niklaus, Long Mai, Jimei Yang, and Feng Liu.
\newblock 3d ken burns effect from a single image.
\newblock {\em ACM TOG}, 38:1--15, 2019.

\bibitem{orts2016holoportation}
Sergio Orts-Escolano, Christoph Rhemann, Sean Fanello, Wayne Chang, Adarsh
  Kowdle, Yury Degtyarev, David Kim, Philip~L. Davidson, Sameh Khamis, Mingsong
  Dou, Vladimir Tankovich, Charles Loop, Qin Cai, Philip~A. Chou, Sarah
  Mennicken, Julien Valentin, Vivek Pradeep, Shenlong Wang, Sing~Bing Kang,
  Pushmeet Kohli, Yuliya Lutchyn, Cem Keskin, and Shahram Izadi.
\newblock Holoportation: Virtual 3d teleportation in real-time.
\newblock In {\em Proceedings of the 29th annual symposium on user interface
  software and technology}, pages 741--754, 2016.

\bibitem{park20103d}
Hyun~Soo Park, Takaaki Shiratori, Iain Matthews, and Yaser Sheikh.
\newblock 3d reconstruction of a moving point from a series of 2d projections.
\newblock In {\em ECCV}, 2010.

\bibitem{park2021nerfies}
Keunhong Park, Utkarsh Sinha, Jonathan~T Barron, Sofien Bouaziz, Dan~B Goldman,
  Steven~M Seitz, and Ricardo Martin-Brualla.
\newblock Nerfies: Deformable neural radiance fields.
\newblock In {\em CVPR}, 2021.

\bibitem{park2021hypernerf}
Keunhong Park, Utkarsh Sinha, Peter Hedman, Jonathan~T Barron, Sofien Bouaziz,
  Dan~B Goldman, Ricardo Martin-Brualla, and Steven~M Seitz.
\newblock Hypernerf: A higher-dimensional representation for topologically
  varying neural radiance fields.
\newblock {\em ACM TOG}, 40, 2021.

\bibitem{penner2017soft}
Eric Penner and Li Zhang.
\newblock Soft 3d reconstruction for view synthesis.
\newblock {\em ACM TOG}, 36:1--11, 2017.

\bibitem{perazzi2016benchmark}
Federico Perazzi, Jordi Pont-Tuset, Brian McWilliams, Luc Van~Gool, Markus
  Gross, and Alexander Sorkine-Hornung.
\newblock A benchmark dataset and evaluation methodology for video object
  segmentation.
\newblock In {\em CVPR}, 2016.

\bibitem{pumarola2021d}
Albert Pumarola, Enric Corona, Gerard Pons-Moll, and Francesc Moreno-Noguer.
\newblock D-nerf: Neural radiance fields for dynamic scenes.
\newblock In {\em CVPR}, 2021.

\bibitem{ranftl2020towards}
Ren{\'e} Ranftl, Katrin Lasinger, David Hafner, Konrad Schindler, and Vladlen
  Koltun.
\newblock Towards robust monocular depth estimation: Mixing datasets for
  zero-shot cross-dataset transfer.
\newblock {\em IEEE TPAMI}, 44, 2020.

\bibitem{riegler2020free}
Gernot Riegler and Vladlen Koltun.
\newblock Free view synthesis.
\newblock In {\em ECCV}, 2020.

\bibitem{riegler2021stable}
Gernot Riegler and Vladlen Koltun.
\newblock Stable view synthesis.
\newblock In {\em CVPR}, 2021.

\bibitem{rosinol2022nerf}
Antoni Rosinol, John~J Leonard, and Luca Carlone.
\newblock Nerf-slam: Real-time dense monocular slam with neural radiance
  fields.
\newblock {\em arXiv preprint arXiv:2210.13641}, 2022.

\bibitem{russell2014video}
Chris Russell, Rui Yu, and Lourdes Agapito.
\newblock Video pop-up: Monocular 3d reconstruction of dynamic scenes.
\newblock In {\em ECCV}, 2014.

\bibitem{schoenberger2016sfm}
Johannes~Lutz Sch\"{o}nberger and Jan-Michael Frahm.
\newblock Structure-from-motion revisited.
\newblock In {\em CVPR}, 2016.

\bibitem{schonberger2016structure}
Johannes~L Schonberger and Jan-Michael Frahm.
\newblock Structure-from-motion revisited.
\newblock In {\em CVPR}, 2016.

\bibitem{shih20203d}
Meng-Li Shih, Shih-Yang Su, Johannes Kopf, and Jia-Bin Huang.
\newblock 3d photography using context-aware layered depth inpainting.
\newblock In {\em CVPR}, 2020.

\bibitem{sinha2022common}
Samarth Sinha, Roman Shapovalov, Jeremy Reizenstein, Ignacio Rocco, Natalia
  Neverova, Andrea Vedaldi, and David Novotny.
\newblock Common pets in 3d: Dynamic new-view synthesis of real-life deformable
  categories.
\newblock {\em arXiv preprint arXiv:2211.03889}, 2022.

\bibitem{sitzmann2019scene}
Vincent Sitzmann, Michael Zollh{\"o}fer, and Gordon Wetzstein.
\newblock Scene representation networks: Continuous 3d-structure-aware neural
  scene representations.
\newblock In {\em NeurIPS}, 2019.

\bibitem{srinivasan2019pushing}
Pratul~P Srinivasan, Richard Tucker, Jonathan~T Barron, Ravi Ramamoorthi, Ren
  Ng, and Noah Snavely.
\newblock Pushing the boundaries of view extrapolation with multiplane images.
\newblock In {\em CVPR}, 2019.

\bibitem{sun2022direct}
Cheng Sun, Min Sun, and Hwann-Tzong Chen.
\newblock Direct voxel grid optimization: Super-fast convergence for radiance
  fields reconstruction.
\newblock In {\em CVPR}, 2022.

\bibitem{teed2020raft}
Zachary Teed and Jia Deng.
\newblock Raft: Recurrent all-pairs field transforms for optical flow.
\newblock In {\em ECCV}, 2020.

\bibitem{teed2021droid}
Zachary Teed and Jia Deng.
\newblock Droid-slam: Deep visual slam for monocular, stereo, and rgb-d
  cameras.
\newblock In {\em NeurIPS}, 2021.

\bibitem{tretschk2021non}
Edgar Tretschk, Ayush Tewari, Vladislav Golyanik, Michael Zollh{\"o}fer,
  Christoph Lassner, and Christian Theobalt.
\newblock Non-rigid neural radiance fields: Reconstruction and novel view
  synthesis of a dynamic scene from monocular video.
\newblock In {\em ICCV}, 2021.

\bibitem{tucker2020single}
Richard Tucker and Noah Snavely.
\newblock Single-view view synthesis with multiplane images.
\newblock In {\em CVPR}, 2020.

\bibitem{wang2021nerf}
Zirui Wang, Shangzhe Wu, Weidi Xie, Min Chen, and Victor~Adrian Prisacariu.
\newblock Nerf--: Neural radiance fields without known camera parameters.
\newblock {\em arXiv preprint arXiv:2102.07064}, 2021.

\bibitem{weng2022humannerf}
Chung-Yi Weng, Brian Curless, Pratul~P Srinivasan, Jonathan~T Barron, and Ira
  Kemelmacher-Shlizerman.
\newblock Humannerf: Free-viewpoint rendering of moving people from monocular
  video.
\newblock In {\em CVPR}, 2022.

\bibitem{wiles2020synsin}
Olivia Wiles, Georgia Gkioxari, Richard Szeliski, and Justin Johnson.
\newblock Synsin: End-to-end view synthesis from a single image.
\newblock In {\em CVPR}, 2020.

\bibitem{xian2021space}
Wenqi Xian, Jia-Bin Huang, Johannes Kopf, and Changil Kim.
\newblock Space-time neural irradiance fields for free-viewpoint video.
\newblock In {\em CVPR}, 2021.

\bibitem{yang2019cubeslam}
Shichao Yang and Sebastian Scherer.
\newblock Cubeslam: Monocular 3-d object slam.
\newblock {\em IEEE Transactions on Robotics}, 2019.

\bibitem{yin2018geonet}
Zhichao Yin and Jianping Shi.
\newblock Geonet: Unsupervised learning of dense depth, optical flow and camera
  pose.
\newblock In {\em CVPR}, 2018.

\bibitem{yoon2020novel}
Jae~Shin Yoon, Kihwan Kim, Orazio Gallo, Hyun~Soo Park, and Jan Kautz.
\newblock Novel view synthesis of dynamic scenes with globally coherent depths
  from a monocular camera.
\newblock In {\em CVPR}, 2020.

\bibitem{yu2018ds}
Chao Yu, Zuxin Liu, Xin-Jun Liu, Fugui Xie, Yi Yang, Qi Wei, and Qiao Fei.
\newblock Ds-slam: A semantic visual slam towards dynamic environments.
\newblock In {\em 2018 IEEE/RSJ International Conference on Intelligent Robots
  and Systems (IROS)}, 2018.

\bibitem{zhang2020vdo}
Jun Zhang, Mina Henein, Robert Mahony, and Viorela Ila.
\newblock Vdo-slam: a visual dynamic object-aware slam system.
\newblock {\em arXiv preprint arXiv:2005.11052}, 2020.

\bibitem{zhang2020nerf++}
Kai Zhang, Gernot Riegler, Noah Snavely, and Vladlen Koltun.
\newblock Nerf++: Analyzing and improving neural radiance fields.
\newblock {\em arXiv preprint arXiv:2010.07492}, 2020.

\bibitem{zhao2022particlesfm}
Wang Zhao, Shaohui Liu, Hengkai Guo, Wenping Wang, and Yong-Jin Liu.
\newblock Particlesfm: Exploiting dense point trajectories for localizing
  moving cameras in the wild.
\newblock In {\em ECCV}, 2022.

\bibitem{zhou2017unsupervised}
Tinghui Zhou, Matthew Brown, Noah Snavely, and David~G Lowe.
\newblock Unsupervised learning of depth and ego-motion from video.
\newblock In {\em CVPR}, 2017.

\bibitem{zitnick2004high}
C~Lawrence Zitnick, Sing~Bing Kang, Matthew Uyttendaele, Simon Winder, and
  Richard Szeliski.
\newblock High-quality video view interpolation using a layered representation.
\newblock {\em ACM TOG}, 23:600--608, 2004.

\end{thebibliography}
}

\end{document}